\documentclass[conference]{IEEEtran}

\IEEEoverridecommandlockouts 

\usepackage{times}
\usepackage{xspace}
\usepackage{bm}

\usepackage{fontawesome5}

\usepackage{color}
\usepackage{colortbl}
\usepackage{xcolor}
\usepackage{subfig}
 
\definecolor{deemph}{gray}{0.6}

\definecolor{baselinecolor}{gray}{.9}
\definecolor{yellow}{RGB}{218,165,32}
\definecolor{lightcyan}{rgb}{0.88, 1.0, 1.0}
\definecolor{lightskyblue}{rgb}{0.53, 0.81, 0.98}
\definecolor{aliceblue}{rgb}{0.94, 0.97, 1.0}
\definecolor{LightSlateBlue}{RGB}{70,130,180}
\definecolor{DeepBlue}{RGB}{65,100,170}
\definecolor{DeepPurple}{RGB}{136,105,160}
\definecolor{LightGreen}{RGB}{59,125,35}
\definecolor{LightRed}{RGB}{234,66,53}
\definecolor{cvprblue}{rgb}{0.21,0.49,0.74}

\usepackage[numbers]{natbib}
\usepackage{multicol}
\usepackage[pagebackref,breaklinks,colorlinks,citecolor=cvprblue,bookmarks=true]{hyperref}
\usepackage{amsmath}
\usepackage{amssymb}
\usepackage{graphicx}
\usepackage{booktabs}
\usepackage{multirow}
\usepackage{xcolor}
\definecolor{darkgreen}{rgb}{0.0,0.5,0.0}
\definecolor{Blue}{HTML}{1976D2}
\definecolor{Purple}{HTML}{6A1B9A}
\definecolor{Green}{HTML}{2E7D32}
\usepackage{caption}
\usepackage{epigraph}
\usepackage{bbm}
\usepackage[ruled,vlined]{algorithm2e}
\usepackage{booktabs}
\usepackage{tabularx}
\usepackage{array}
\usepackage{wrapfig}
\SetKwInput{Input}{Input}
\SetKwInput{Output}{Output}
\definecolor{DeepBlue}{RGB}{0,82,147}

\newcommand{\name}{$\chi_0$\xspace}



\hypersetup{
 colorlinks=true,
 linkcolor=red,
 citecolor=cvprblue,
 filecolor=magenta, 
 }

\begin{document}

\title{\name{}: Resource-Aware Robust Manipulation via Taming Distributional Inconsistencies}

\author{
\authorblockN{
Checheng Yu,
Chonghao Sima,
Gangcheng Jiang,
Hai Zhang,
Haoguang Mai, \\
Hongyang Li, 
Huijie Wang,
Jin Chen,
Kaiyang Wu,
Li Chen,
Lirui Zhao, \\
Modi Shi, 
Ping Luo,
Qingwen Bu,
Shijia Peng,
Tianyu Li,
Yibo Yuan
}
\smallskip
\authorblockA{
\textbf{Kinetix AI} \\
{\small 
\faGithub\ Code: \texttt{\url{https://github.com/OpenDriveLab/kai0}}} \\
{\small 
\faFile*\ Blog: \texttt{\url{https://mmlab.hk/research/kai0}}}
}
\thanks{Authors listed alphabetically by first name.
}
\thanks{Contact: \texttt{chonghaosima@connect.hku.hk}}
}
\makeatletter
\let\@oldmaketitle\@maketitle%
\renewcommand{\@maketitle}{\@oldmaketitle %
  \centering
  \setcounter{figure}{0}%
  \includegraphics[width=0.99\textwidth]{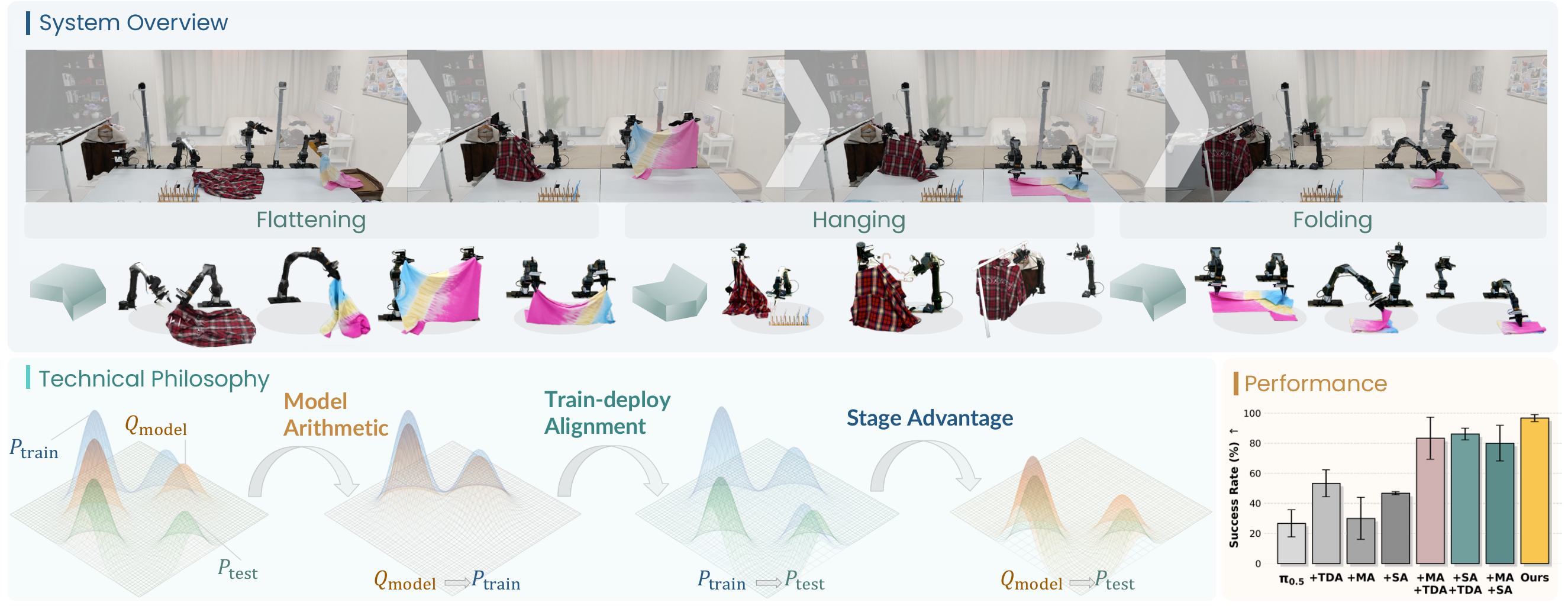}
    \captionof{figure}{
    \textbf{Top: System overview.} A robot teamwork system with two dual-arm ALOHA robots performing 
    long-horizon collaborative garment manipulation, including flattening, folding and hanging.
    \textbf{Bottom: Technical philosophy and performance.} Distributional inconsistencies are inherent to robot learning ($P_\text{train}$: expert demonstrations; $Q_\text{model}$: policy inductive bias; $P_\text{test}$: deployment trajectories). \name{} systematically resolves these pairwise mismatches: Model Arithmetic aligns $Q_\text{model}$ with $P_\text{train}$; Train-Deploy Alignment bridges $P_\text{train}$ and $P_\text{test}$; and Stage Advantage optimizes $Q_\text{model}$ for $P_\text{test}$. The
    contributions of these modules collectively enable \name{} to surpass the baseline $\pi_{0.5}$~\cite{black2025pi05} in terms of success rate by approximately 250\%. 
    }
  \label{fig:teaser}
}
\makeatother

\maketitle

\begin{abstract}
High-reliability long-horizon robotic manipulation
has traditionally relied on
large-scale
data and compute
to understand complex real-world dynamics.
However, 
we identify that 
the primary bottleneck to real-world robustness is not resource scale alone, but the 
distributional shift
among the human demonstration distribution, the inductive bias learned by the policy, and the test-time execution distribution—a systematic inconsistency that causes compounding errors in multi-stage tasks.
To mitigate these inconsistencies, we propose \name{}, a resource-efficient framework with effective modules
designated to achieve production-level robustness in robotic manipulation.
Our approach builds off three technical pillars:
(i) Model Arithmetic, a weight-space merging strategy that efficiently soaks up diverse distributions of different demonstrations, varying from object appearance to state variations;
(ii)
Stage Advantage, a stage-aware advantage estimator that provides stable, dense progress signals,
overcoming the numerical instability of prior non-stage approaches;
and (iii) Train-Deploy Alignment, 
which bridges the distribution gap via spatio-temporal augmentation, heuristic DAgger corrections, and temporal chunk-wise smoothing.
\name{} enables two sets of dual-arm 
robots to collaboratively orchestrate long-horizon garment manipulation, spanning tasks from
flattening, folding, to hanging different clothes.
Our method exhibits high-reliability autonomy; we 
are
able to run the system
from arbitrary initial state for consecutive 24 hours non-stop.
Experiments validate that \name{} surpasses the state-of-the-art
$\pi_{0.5}$ in success rate by nearly 250\%, with only 20-hour data and $8$ A100 GPUs. 
Code, data and models will be released to facilitate the community.
\end{abstract}

\IEEEpeerreviewmaketitle

\section{Introduction}
\label{sec:introduction}
Achieving production-ready robustness stands as the grand challenge of modern robotics.
While autonomous vehicles, as a specific form of navigation robot, have successfully demonstrated operational viability in complex urban environments~\cite{Kusano2025waymocrash,tesla2025fsd,waymo2025blog,hu2023uniad,sima2024drivelm,sima2025centaur}, replicating this level of reliability for robotic manipulation in unstructured settings and garnering a significant degree of human trust still remain as an open challenge.
Achieving such robustness requires a policy to operate within a vast search space, handling the tremendous environmental variability inherent to the physical world.
Therefore, the prevailing industrial paradigm has pivoted toward a scaling approach, leveraging massive computational resources to scale foundation models~\cite{black2024pi0,black2025pi05,black2025pistar,generalist2025gen0,sunday2025act1,zheng2025xvla,beingbeyond2026beingh05}.
However, while architectural evolution and resource scaling are essential, we argue that the decisive factor in robust policy execution is not scale alone.
Our key insight is that within the expansive search space in real world for the policy, the ``hidden devil" hindering robustness is \textit{the inconsistency among the distributions governing the three pillars of the robot learning: data collection, model training and policy deployment}.
These inconsistencies are not evidently reflected in success rates, but rather in the smoothness of execution, system throughput, and the retry cost required for successful task completion~\cite{chen2025benchmarking,bu2025agibot,shi2025diversity,black2025pistar,jiang2025behavior}.
We encapsulate the robot learning pipeline into three distinct distributions spanning the entire research cycle to formalize further analysis -- 
\bm{$P_\text{train}$}, the distribution of human expert demonstrations used for training an imitation policy; 
\bm{$Q_\text{model}$}, the distribution of the inductive bias learned by the policy. A state-to-plausible-actions mapper; 
\bm{$P_\text{test}$}, the distribution of executed action trajectories during real-robot deployment. It differs from the policy output action with certain delay and physical limitation;

Massive real-world deployment of learned policies reveals three systematic inconsistencies within this regime, as shown in 
Figure~\ref{fig:teaser}. First, due to the extreme high dimensionality of the task, $P_\text{train}$
is inherently sparse relative to the full solution manifold,
resulting in a $Q_\text{model}$ that is heavily biased toward the limited training distribution. 
Second, the latency between model inference ($Q_\text{model}$) and control-level execution ($P_\text{test}$) introduces a temporal mismatch, rendering theoretically optimal plans suboptimal during inference~\cite{black2025rtc,tang2025vlash}.
Third, despite frequent failures during inference, the policy lacks failure recovery ability; even when encountering states within 
$P_\text{train}$, minor perturbations in $P_\text{test}$ can trigger catastrophic divergence from which the system cannot recover~\cite{hu2025rac,pan2026sop,black2025pistar}.
Prior literature addresses these observed inconsistencies via strategies such as dataset scaling~\cite{wiles2022fine,higgins2017beta,radford2021learning}, heuristic or learned augmentation~\cite{wiles2022fine,cubuk2020randaugment,cubuk2019autoaugment}, and adaptive learning~\cite{wiles2022fine,liu2021just,schneider2020improving}.
However, the off-the-shelf application of these general-purpose methods to robotic manipulation is hindered by domain-specific constraints: the prohibitive cost of collecting expert demonstrations, significant inference-to-execution latency, and the computational burden of training large-scale models.
To bridge this gap,
we propose \name{}, 
a holistic framework designed to systematically resolve these distributional misalignments within the constraints of physical robotics.
Our approach builds off three technical pillars
that ease the inconsistencies one by one: 
\begin{enumerate}
    \item[(a)] \textbf{Model Arithmetic (MA):} MA is to align varying data subsets ($P_\text{train}$) with the policy's inductive bias ($Q_\text{model}$). This approach enables the policy to efficiently soak up diverse $P_\text{train}$ distributions by simply merging weight of checkpoints trained on different $P_\text{train}$.
    As shown in Figure~\ref{fig:teaser}, MA enables $Q_\text{model}$ to capture modes within $P_\text{train}$ that were previously omitted.

    \item[(b)] \textbf{Stage Advantage (SA):} To optimize action sampling ($Q_\text{model}$) under the novel deployment environment ($P_\text{test}$), SA decomposes long-horizon tasks into semantic sub-goals (referred as stages), providing stable, stage-aware reward signals for advantage-weighted behavior cloning~\cite{peng2019advantage}.
    Figure~\ref{fig:teaser} demonstrates the idea that SA allows $Q_{models}$ to sample actions in the mode that is closer to $P_\text{test}$.
    Furthermore, SA mitigates the numerical instability inherent in prior non-stage methods like $\pi^{*}_{0.6}$~\cite{black2025pistar} via frame-wise reward modeling.

    \item[(c)] \textbf{Train-Deploy-Alignment (TDA):} TDA 
    expands
    $P_\text{train}$ toward $P_\text{test}$ via heuristic DAgger and spatio-temporal augmentation, ensuring robustness against real-world distributional drift.
    We further propose temporal chunk-wise smoothing to mitigate inference-actuation latency and enhance real-time control stability, surpassing RTC-only method~\cite{black2025rtc,tang2025vlash} in terms of policy throughput and retry cost.
    We illustrate this idea in Figure~\ref{fig:teaser}, that $P_\text{train}$ achieves improved coverage of the modes within $P_\text{test}$.
\end{enumerate}

We evaluate \name{} on collaborative long-horizon garment manipulation tasks such as flattening, folding, and hanging different clothes, as contact-rich, deformable dynamics of clothes and recovery from arbitrary cloth states magnify the distributional shifts aforementioned.
Training with just 20 hours of demonstrations on 8$\times$A100 GPUs, \name{} outperforms the open-source $\pi_{0.5}$ baseline by nearly \textbf{250\%} in success rate.
Our extensive experiments empirically align with our insight:
\begin{enumerate}
    \item MA provides a resource-efficient mechanism to enhance policy performance across nearly all metrics; we find that validation loss on DAgger data serves as an effective heuristic for weighting several checkpoints;
    \item In TDA, the DAgger data proves critical for maximizing success rates, albeit at the expense of increased retry costs. This trade-off is consistent with intuition: DAgger samples are most valuable in recovery scenarios, implying that higher retry frequency positively correlates with ultimate task success.
    We also observe that spatio-temporal augmentation is effective only when paired with control optimization, where our proposed temporal chunk-wise smoothing operates orthogonally to RTC~\cite{black2025rtc};
    \item Regarding SA, our stage-aware 2-timestep advantage signal exhibits superior numerical stability compared to $\pi^{*}_{0.6}$-style advantage training~\cite{black2025pistar}, a stability that empirically translates to improved overall performance.
\end{enumerate}
All these contributions sustain a livestream of autonomous operation in a 24-hour real-robot stress test, provided in the Appendix.
We will release data, code, and model weights
to facilitate the community.
\section{Related Work}
\label{sec:related_work}
\begin{figure*}[t]
  \centering
  \includegraphics[width=0.99\linewidth]{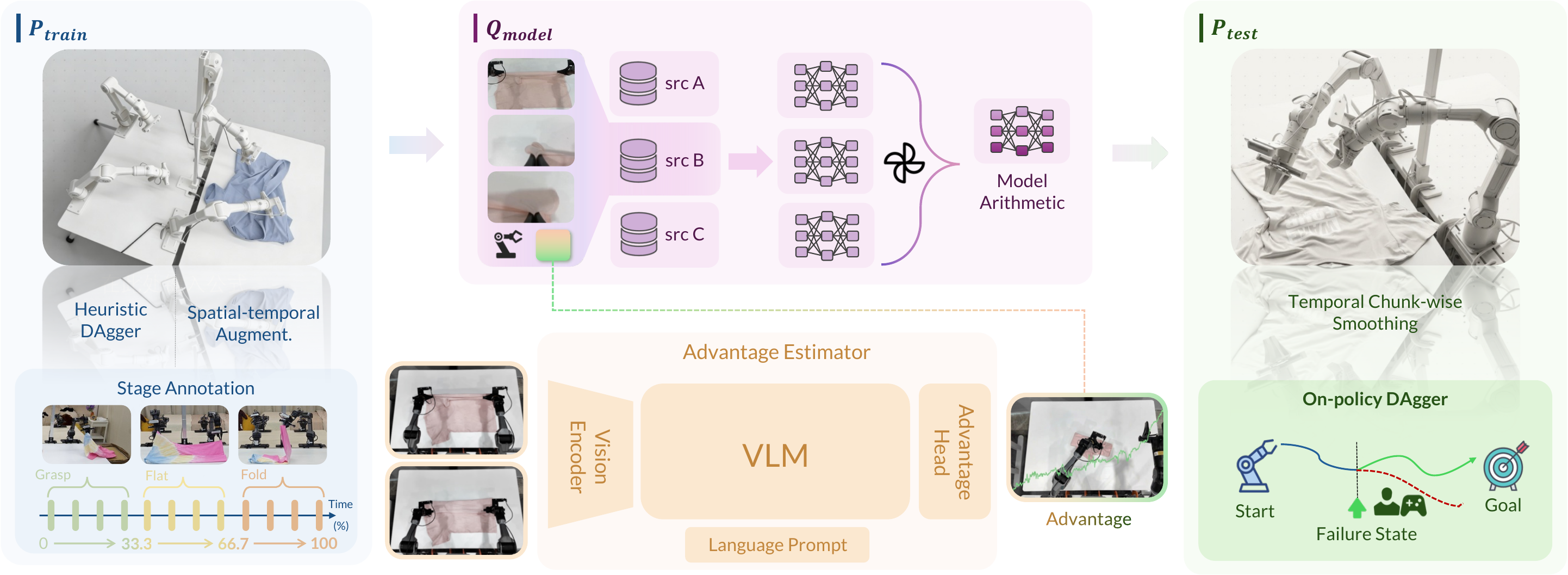}
  \caption{\textbf{Pipeline of \name{}.} Our framework addresses distributional inconsistencies across three stages. \textbf{(Left)} $P_{\text{train}}$: heuristic DAgger and spatio-temporal augmentation expand training coverage, with Stage Annotation for advantage estimation; \textbf{(Middle)} $Q_{\text{model}}$: Model Arithmetic merges complementary policies in weight space, guided by stage-aware advantage; \textbf{(Right)} $P_{\text{test}}$: temporal chunk-wise smoothing ensures execution accuracy, while on-policy DAgger enables closed-loop refinement.}
  \label{fig:pipeline}
  \vspace{-1em}
\end{figure*}

\subsection{Imitation Learning and Policy Deployment in Real-world}
Imitation learning has become a dominant paradigm for robot manipulation, scaling from lightweight transformer-based policies~\cite{zhao2023aloha, chi2024umi, chi2023diffusion,zeng2024learning} to foundation models trained on massive robot demonstrations~\cite{brohan2022rt1, brohan2023rt2, kim2024openvla, team2024octo, bu2025univla, black2025pi05, black2025pistar}. Among these, \(\pi\) series~\cite{black2024pi0, black2025pi05, black2025pistar} stand out for strong generalization via large-scale pretraining dataset.
Yet the collection of such datasets~\cite{khazatsky2024droid,padalkar2023open,bu2025agibot,jiang2025galaxea,walke2023bridgedata} demands substantial resources.
For data efficiency, prior works explore DAgger-style aggregation~\cite{ross2011dagger, kelly2018hgdagger, hu2025rac, black2025pistar} and augmentation~\cite{laskin2020reinforcement, kostrikov2020image, yu2023scaling, li2025gr}. However, DAgger requires real-time human supervision during policy rollouts~\cite{ross2011dagger}, making data collection still time-consuming.
Besides, real-world deployment introduces unique challenge to these polices: inference-control latency causes mismatch between model outputs and physical execution. Prior works try to mitigate this through execution-side optimizations~\cite{zhao2023aloha, black2025rtc, tang2025vlash}, yet introduce additional inference overhead.

While existing methods address individual phases rather than jointly enforcing distributional consistency across the robot learning cycle: data collection, model training and deployment ~\cite{chen2025retaining,tajwar2024preference},
we formalize this challenge through \(P_{\text{train}}\), \(Q_{\text{model}}\), \(P_{\text{test}}\) and propose \name{} to comprehensively align them.

\subsection{Model Merging and Weight Interpolation}

Model merging has emerged as an efficient strategy for consolidating knowledge from multiple neural networks.
Initial work in computer vision and natural language processing demonstrated that interpolating weights between models across hyperparameter perturbation checkpoints~\cite{wortsman2022model} or across models fine-tuned on different tasks~\cite{wortsman2022model,ilharco2023editing, Yadav2023TIESMergingRI} can improve generalization and robustness. 
These techniques have recently been extended to large language models \cite{Akiba2024EvolutionaryOO,Yang2024ModelMI,Team2025KimiKS}, planning \cite{Lee2025InteractionMergedMP}, and robot learning \cite{Wang2023RobotFL}.
These methods often rely on in-distribution metrics for merging strategy selection, which may not account for the narrow distribution shifts common in complex manipulation. 
Parallel to our work, RETAIN \cite{Yadav2025RobustFO} applies model merging to adapt VLA policies, improving target-task out-of-distribution (OOD) generalization while better retaining generalist skills.
Our work, Model Arithmetic (MA), specifically mitigates the model bias caused by incomplete training coverage with limited expert demonstration. 
We introduce a novel validation protocol using OOD data---specifically recovery trajectories collected via DAgger---to ensure the merged policy generalizes to unseen states. 
Furthermore, we provide a comparative analysis of multiple merging strategies, including uniform weighting, inverse loss weighting, gradient decent, and greedy search, to identify the most effective synthesis for mitigating training data bias.

\subsection{Advantage Estimation for Long-Horizon Tasks}

Prior work has explored conditioning policies on rewards, values, and advantages to guide action selection in long-horizon tasks\cite{schmidhuber2019reinforcement, emmons2022rvs, zheng2022online, wu2023elastic, kuba2023advantage}. This includes advantage-weighted objectives such as advantage-weighted regression (AWR), which biases behavior cloning toward higher-advantage actions\cite{peng2019advantage}. Building on these ideas, $\pi^{*}_{0.6}$ trains a distributional value model to estimate state-action advantages and uses them for advantage-conditioned VLA training \cite{black2025pistar}.

A key limitation in practice is numerical instability: advantages computed from value differences can be noisy and high-variance, especially under long-horizon real-world dynamics.
\textit{Stage Advantage} addresses this by directly predicting advantage from paired observations and conditioning the signal on semantic stages, yielding a smoother and more stable supervision signal that can be discretized into a binary optimality indicator for policy learning \cite{li2025self, black2025pistar}.
\section{Methodology}

\label{sec:methodology}

\subsection{Preliminary and Problem Setup}
\label{subsec:preliminary}
To formalize the distributional framework introduced in Sec.~\ref{sec:introduction},
we consider a finite-horizon Markov Decision Process (MDP)
with state space $\mathcal{S}$, action space $\mathcal{A}$ and horizon $H$. 
A trajectory $\tau = (s_0, a_0, s_1, a_1, \dots, s_H)$ evolves under real-world dynamics $s_{t+1} \sim T(\cdot \mid s_t, a_t, \xi)$ under a fixed environment parameterization $\xi$, with initial-state distribution $\mu(s_0)$.
Accordingly, the trajectory distribution induced by a stochastic policy $\pi(a \mid s;\phi)$ could be defined as~\cite{schulman2015trust, levine2018reinforcement}:
{
    \begin{equation}
    \label{eq: traj_distrib}
        \scalebox{0.95}{ $P_{\pi}(\tau) = \mu(s_0) \prod_{t=0}^{H-1} \pi(a_t \mid s_t;\phi) \, T(s_{t+1}\mid s_t, a_t, \xi),$ }
    \end{equation}
}
abbreviated as $P$ when the policy $\pi$ is unambiguous.

We first define $P_{\text{real}}$ as the distribution over successful trajectories that complete the tasks. 
Let $\mathcal{G}$ denote the set of successful trajectories, such that
\(P_{\text{real}}(\tau) \propto \mathbf{1}\{\tau \in \mathcal{G}\}, \)
characterizing the manifold of all valid action sequences achieving task completion under real-world dynamics.
With \(P_{\text{real}}\) we now formalize the three distributions underlying our alignment objective. 
The training distribution \(P_{\text{train}}\) is induced by human expert demonstrations on real robots.
Given \(P_{\text{train}}\), let 
\(\mathcal{D} = {\{\tau^{(i)}\}_{i=1}^N}\)
denote the set of demonstrations.
The model \(Q_{\text{model}}\triangleq \pi(a\mid s;\hat\phi)\) is learned by maximizing \(\sum_{t}\log \pi(a_t\mid s_t;\phi)\) over the demonstration set \(\mathcal{D}\)~\cite{ross2011dagger, osa2018algorithmic}.
Finally, real-robot execution produces \(P_{\text{test}}\) by composing \(Q_{\text{model}}\) with an inference operator \(I(\tilde a_t\mid a_t,s_t)\):
{
\begin{equation}
\scalebox{0.95}{$P_{\text{test}}(\tau)
= \mu(s_0)\prod_{t=0}^{H-1}
\pi(a_t\mid s_t;\hat\phi)\,
\tilde T(s_{t+1}\mid s_t,a_t,\xi),$}
\end{equation}
}
where
\(
\tilde T(s_{t+1}\mid s_t,a_t,\xi)
\triangleq \mathbb{E}_{\tilde a\sim I(\cdot\mid a_t,s_t)}
\big[\,T(s_{t+1}\mid s_t,\tilde a_t,\xi)\,\big]
\), and \(\tilde a_t\) denotes the actually executed action.

As discussed in Sec.~\ref{sec:introduction}, real-world deployment reveals three systematic inconsistencies among these distributions across different phases. 
We categorize them as: 
\textbf{(i) Coverage Deficiency}---\(P_{\text{train}}\) undersamples the high-dimensional manifold \(P_{\text{real}}\), biasing \(Q_{\text{model}}\) toward limited training support; 
\textbf{(ii) Temporal Mismatch}---long-horizon tasks induce visually similar but semantically distinct states across stages, causing \(Q_{\text{model}}\) to misapply temporal knowledge, while inference-control latency also leads to execution-level temporal mismatch, all of which are reflected as failure or staying still in $P_\text{test}$;
\textbf{(iii) Failure Cascade}---absence of recovery behaviors in \(P_{\text{train}}\) leaves the policy unable to self-correct from perturbations in $P_\text{test}$. 
Our method tries to address each inconsistency through targeted alignment strategies, detailed in the following sections.

\subsection{Pipeline of \name{} system}

To coherently address the three distributional inconsistencies identified in Sec.~\ref{subsec:preliminary}, we propose \name{}
, a resource-efficient framework integrating three 
complementary
technical pillars across the robot learning cycle. 
Model Arithmetic (Sec.~\ref{subsec:model_arithmetic}) expands policy coverage by merging models trained on complementary data subsets in weight space;
Stage Advantage (Sec.~\ref{subsec:stage_advantage}) addresses temporal mismatch at policy learning phase through stage-aware advantage estimation for stable long-horizon supervision; 
Train-Deploy-Alignment (Sec.~\ref{subsec:mode_consistency}) closes the loop between deployment and training through inference optimization and complementary data augmentation. 
Figure~\ref{fig:pipeline} illustrates the complete pipeline. 

\subsection{Model Arithmetic}
\label{subsec:model_arithmetic}

\begin{figure}[t!]
  \centering
  \includegraphics[width=\columnwidth]{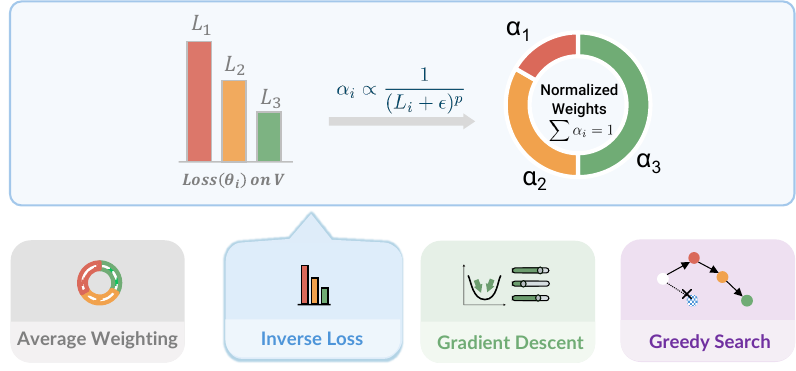}
  \caption{\textbf{Souping strategies in Model Arithmetic.} Policies trained on separate subsets are merged via weighted interpolation. \textbf{(Top)} Inverse Loss assigns higher coefficients to models with lower validation loss. \textbf{(Bottom)} Other strategies.}
  \label{fig:arithmetic}
  \vspace{-1em}
\end{figure}

Limited expert demonstrations in the initial data collection phase lead to coverage deficiency in \(P_{\text{train}}\), which further biases the learned policies toward narrow manipulation patterns.
To mitigate this, a straightforward solution is to scale up expert demonstrations until \(P_{\text{train}}\) sufficiently approximates \(P_{\text{real}}\). 
However, it is prohibitively expensive for garment manipulation: each collection cycle demands extensive operation time. 
Consequently, it raises a fundamental question:
How can we efficiently mitigate the model bias without scaling data?

We propose \textbf{Model Arithmetic (MA)}, a weight-space merging strategy that combines policies trained on complementary data subsets, guided by validation-based optimization. 
Unlike Mixture-of-Experts (MoE), which requires explicit router mechanisms and complex training design~\cite{shazeer2017outrageously, fedus2022switch}, or model ensembling that combines model outputs~\cite{lakshminarayanan2017simple}, MA directly merges parameters to synthesize a unified policy.
Formally, given collected data subsets $\{D_1, D_2, \ldots, D_n\}$ sampled from \(P_{\text{train}}\), MA trains policies $\{\theta_1, \theta_2, \ldots, \theta_n\}$ independently on these subsets and synthesizes their model weights via interpolation:
\(\theta_{\text{merged}} = \sum_{i=1}^{n} \alpha_i \theta_i, \quad s.t. \alpha_i \ge 0,\ \sum_{i=1}^{n} \alpha_i = 1.\)
$\{\alpha_i\}$ is optimized by minimizing a held-out loss 
over
validation 
split.
\(\theta_{\text{merged}}\) serves as the final \(Q_{\text{model}}\) for deployment.

MA starts with randomly partitioning the training dataset \(\mathcal{D}\) into non-overlapping subsets \(\{\mathcal{D}_1, \mathcal{D}_2, \dots, \mathcal{D}_n\}\) and training separate policies on each. Due to limited coverage per subset, these policies naturally converge to distinct regions of the solution manifold. The key challenge then becomes how to optimally merge these policies.
In practice, the critical design choices lies in validation set selection. We strategically construct a validation set that is out-of-distribution (OOD) with respect to all training subsets (in-domain), ensuring unbiased evaluation of merged policy. Specifically, we use trajectories collected via DAgger~\cite{ross2011dagger, kelly2018hgdagger}, from models trained on individual subsets, as these recovery behaviors are naturally absent from any original training data.
Based on this validation set, we implement and ablate four souping strategies to obtain the final \(\theta_{\textbf{merged}}\)---average weighting, inverse loss, gradient descent and greedy search~\cite{wortsman2022model}---as illustrated in Figure~\ref{fig:arithmetic}.

Through validation-guided weight-space synthesis, MA effectively combines diverse unimodal policies into a unified multi-modal policy, mitigating the bias of \(Q_{\text{model}}\) induced by coverage deficiency without additional data collection. 
\subsection{Stage Advantage}
\label{subsec:stage_advantage}
\begin{figure}[t!]
  \centering
  \includegraphics[width=0.96\columnwidth]{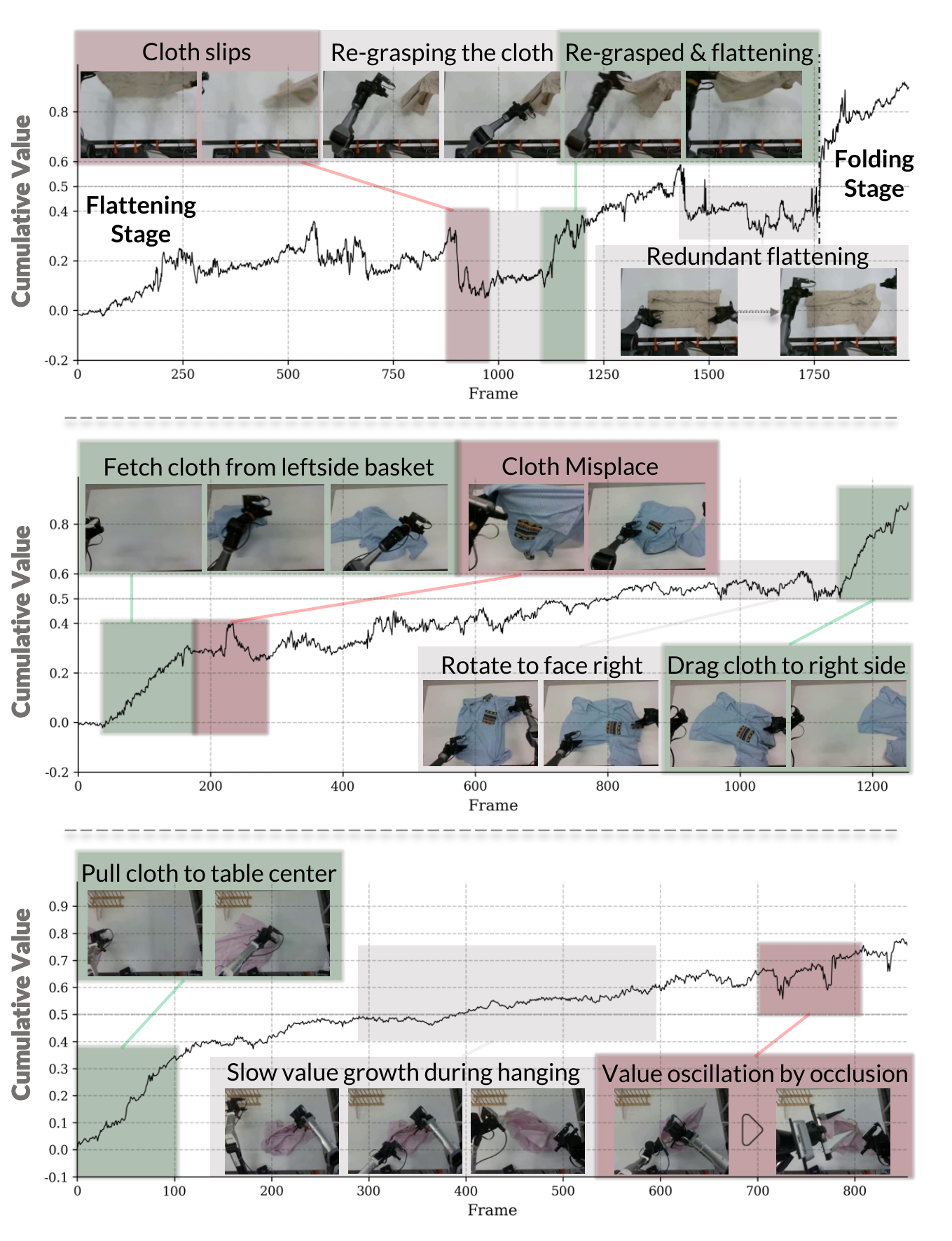}
  \caption{
  \textbf{Cumulative value based on SA.} Red/green stands for negative/positive. \textbf{Top}: Task A shows slip fails and recovery; \textbf{Middle}: Task B shows fetching and cloth misplace; \textbf{Bottom}: Task C shows pull-over and visual occlusion.
  }
  \label{fig:advantage}
  \vspace{-1.5em}
\end{figure}

While Model Arithmetic efficiently mitigates \(Q_{\text{model}}\) bias, the resulting policy still struggles with long-horizon execution in \(P_{\text{test}}\) due to temporal mismatch: 
visually similar states across task stages cause the policy to misapply behaviors, leading to compounding errors and task failure over long horizons. The stage ambiguity calls for explicit progress signals that can disambiguate action quality within the context of task progress~\cite{li2025self}. This raises a key question: how to provide stable and accurate progress signals during long-horizon execution?

Prior approach~\cite{black2025pistar} uses advantage as the progress signal, combining with advantage-weighted regression \cite{wu2023elastic, kuba2023advantage} to train policies with advantage-weighted training samples.
It obtains advantage implicitly as \(A(s, a) = V(s') - V(s),\) taking the difference between independently predicted progress value. 
Nevertheless, this formulation would amplify frame-wise   estimation noise, yielding high-variance training signals. 
Moreover, estimating global task progress without stage awareness causes \(V(s)\) to exhibit multi-valued predictions for multi-stage tasks, further degrading advantage quality.

To obtain a stable and accurate advantage signal for model training,
we take a more straightforward route by treating advantage as a \textbf{direct modeling target}:
\(
    A(s, a) = f_\theta(s, s'),
\) 
where \(f_{\theta}\) predicts relative progress from $s$ to $s'$.
This recasts advantage estimation to a single prediction, avoiding error compounding and yielding a smoother, more reliable state-to-state supervision signal.
In practice, we use a VLM-based architecture that takes pairwise image inputs as the advantage estimator, as shown in Figure~\ref{fig:pipeline}.
To avoid overfitting to a fixed temporal discretization, we construct training pairs by randomly sampling a time span $\Delta$ and setting $s' = s_{t+\Delta}$.

To further resolve the multi-valued ambiguity in progress estimation over long-horizon tasks, \textbf{Stage Advantage} decomposes the task into a sequence of semantic stages, each corresponding to a meaningful sub-goal.
Instead of evaluating actions under full task horizon, we estimate whether each action advances the current stage, providing a stage-aware progress signal:
\(A_{\text{stage}}(s, a, g) = f_\theta(s, s' | g)\).
Practically we use manually annotated stage labels to represent the stage as a normalized scalar $g \in \{0,\tfrac{1}{S},\ldots,\tfrac{S-1}{S}\}$, $S$ is the number of stages as Figure~\ref{fig:pipeline} shows.
Figure~\ref{fig:advantage} shows cumulative value based on stage advantage for tasks defined in Sec.~\ref{sec:exp_setup}.

Following~\cite{li2025self, black2025pistar}, we threshold the continuous advantage predictions into a binary optimality indicator $I=\mathbbm{1}[A_{\text{stage}}>\epsilon]$, where \(\epsilon\) is a threshold that separates progress from non-progress. This enables stable advantage-weighted policy learning that upweights high-quality data from \(P_\text{train}\) while mitigating the temporal mismatch between \(P_{\text{train}}\) and \(Q_{\text{model}}\).

\subsection{Train-Deploy-Alignment}
\label{subsec:mode_consistency}

\begin{figure}[t!]
  \centering
  \includegraphics[width=\columnwidth]{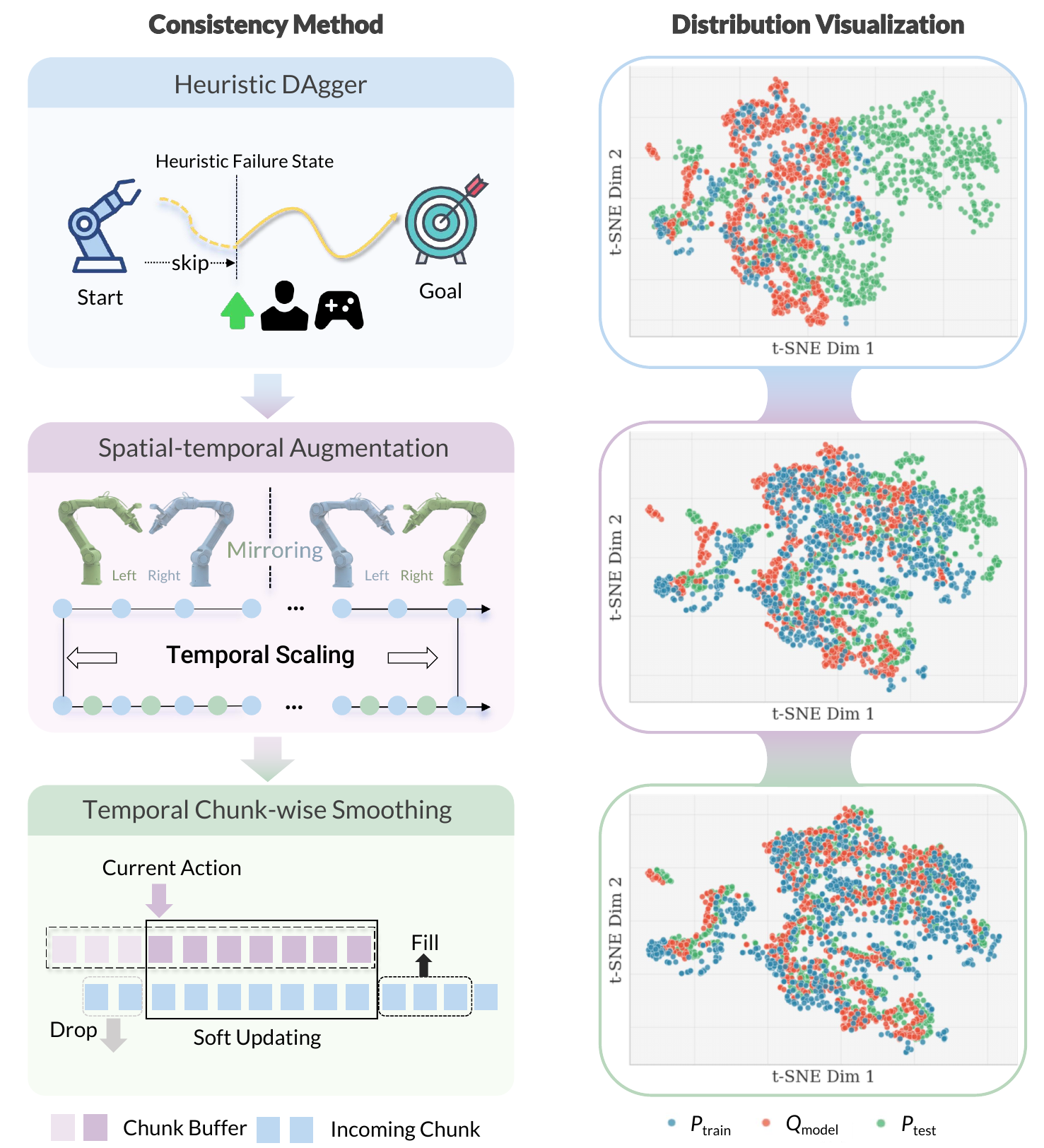}
  \caption{
  \textbf{Train-Deploy-Alignment strategies and T-SNE visualizations.} \textbf{Left}: three complementary strategies for distribution alignment.
  \textbf{Right}: T-SNE visualizations showing progressive distribution alignment as each strategy is applied.
  }
  \label{fig:consistency}
  \vspace{-1.5em}
\end{figure}

Despite robust policies with long-horizon planning ability, real-world deployment introduces new inconsistencies between \(Q_{\text{model}}\) and \(P_{\text{test}}\). 
Inference-control latency causes misplaced action execution and drift error accumulation, especially for action-chunking policies that output action chunks: the gap between model inference and chunk execution breaks temporal continuity across consecutive chunks, leading to abrupt transitions and degraded manipulation stability. 
Prior work addresses this through inference-time chunk interpolation~\cite{zhao2023aloha, black2025rtc, tang2025vlash}. Additionally, we adopt a \textbf{temporal chunk-wise smoothing} to ensure coherent action executions in deployment phase as the bottom part of Figure~\ref{fig:consistency}.

\begin{figure}[t!]
  \centering
  \includegraphics[width=\columnwidth]{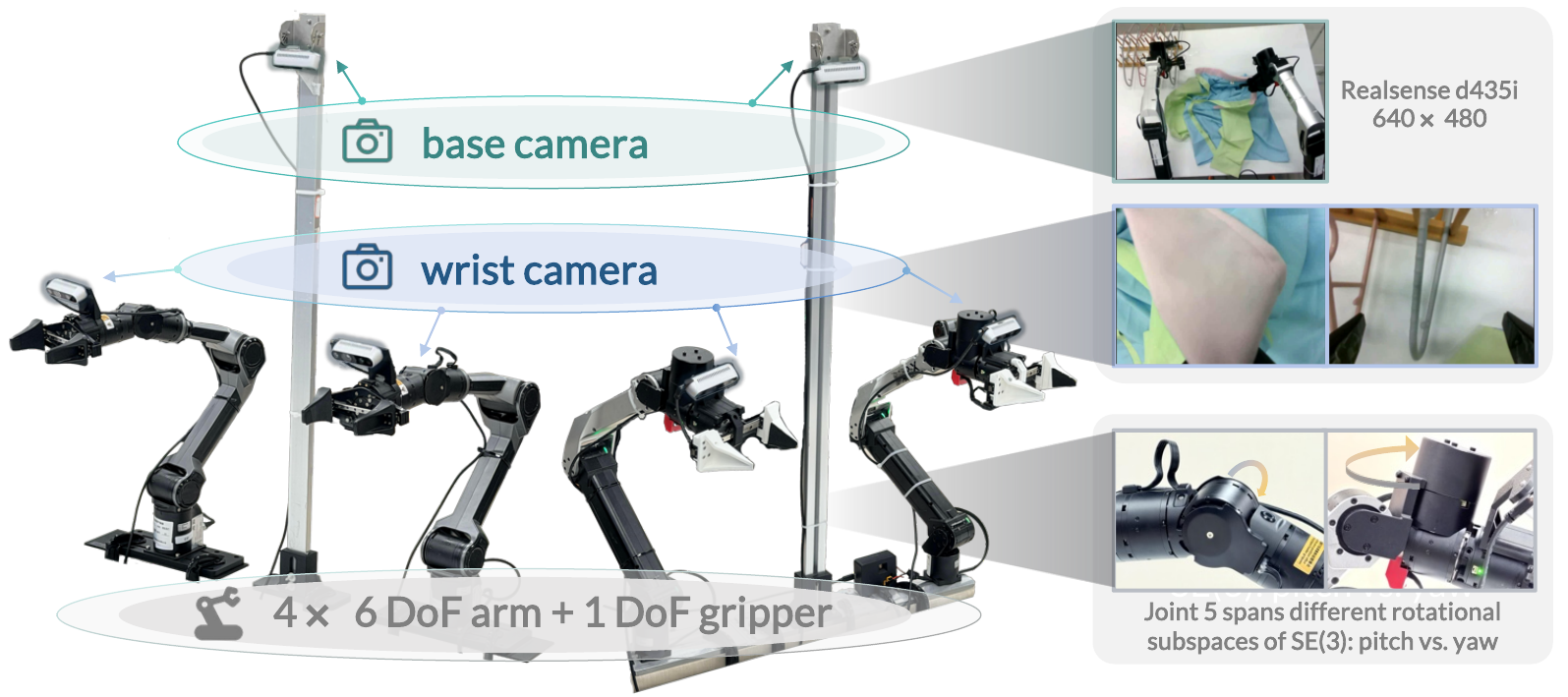}
  \caption{
  \textbf{Robot setup} of our collaborative dual-arm system.
  }
  \label{fig:robot_setup}
  \vspace{-1.5em}
\end{figure}

Mathematically, let \(a^{old}\) denote the current action buffer containing residual commands from previous inference cycle, and \(a^{new}\) the newly predicted action chunk. We maintain a consumption index \(k\) tracking executed actions in current action buffer, 
a drop threshold \(d_{\text{max}}\) to discard stale commands caused by inference latency,
and a minimum overlap length \(m_{\text{min}}\) to ensure stable interpolation.
Based on these, we present the detailed smoothing procedure in Algorithm~\ref{alg:chunk_smoothing}.

{
    \vspace{-10pt}
    \begin{algorithm}[hb]
    \caption{Temporal Chunk-wise Smoothing}
    \label{alg:chunk_smoothing}
    {
        \footnotesize
        \SetAlFnt{\footnotesize}
        \SetKwInOut{Input}{Input}
        \SetKwInOut{Output}{Output}
        \Input{current buffer $\mathbf{a}^{\text{old}}$, index $k$, new chunk $\mathbf{a}^{\text{new}}$, $d_{\max}$, $m_{\min}$}
        \Output{updated buffer $\mathbf{a}^{\text{buf}}$, reset index $k\leftarrow 0$}
        $d \leftarrow \min(k,d_{\max})$\tcp*{compute drop amount}
        \If{$d \ge |\mathbf{a}^{\text{new}}|$}{
          \Return $\mathbf{a}^{\text{old}},k$\tcp*{ignore update}
        }
        $\mathbf{a}^{\text{new}}_{\text{rem}} \leftarrow (a^{\text{new}}_{d},\dots)$\tcp*{remaining new commands}
        \If{$|\mathbf{a}^{\text{old}}|<m_{\min}$}{
          pad $\mathbf{a}^{\text{old}}$ by repeating last command\;
        }
        $L \leftarrow \min(|\mathbf{a}^{\text{old}}|,|\mathbf{a}^{\text{new}}_{\text{rem}}|)$\tcp*{overlap length}
        \For{$i\leftarrow 0$ \KwTo $L-1$}{
          $w_i \leftarrow 1 - i/\max(L-1,1)$\;
          $\tilde a_i \leftarrow w_i a^{\text{old}}_i + (1-w_i) a^{\text{new}}_{\text{rem},i}$
        }
        $\mathbf{a}^{\text{buf}} \leftarrow (\tilde a_0,\dots,\tilde a_{L-1})\Vert(\text{suffix of }\mathbf{a}^{\text{new}}_{\text{rem}})$\;
        $k\leftarrow 0$\tcp*{reset consumption index}
        \Return $\mathbf{a}^{\text{buf}},k$
    }
    
    \end{algorithm}
    \vspace{-10pt}
}

With a robust policy and reliable deployment pipeline established, a natural question arises: 
\textit{can we leverage rollout experience from \(P_{\text{test}}\) to expand 
\(P_{\text{train}}\) without extended data collection effort?} Recall that static demonstrations lack recovery behaviors, leaving the policy vulnerable to failure cascades. We address this final inconsistency by closing the loop between deployment and training through two complementary strategies.
\textbf{1)} On-policy DAgger~\cite{ross2011dagger, kelly2018hgdagger} expands 
\(P_{\text{train}}\)
toward failure-adjacent regions but is time-consuming, as it requires waiting for natural failures during policy rollouts. We propose a \textbf{Heuristic DAgger} variant that directly initializes the system in manually designed failure states (e.g., misaligned grasps, partial drops) and collects recovery demonstrations, front-loading failure experience into data collection. 
\textbf{2)} To further diversify \(P_{\text{train}}\) at zero robot time, we apply \textbf{Spatio-temporal Augmentations}: horizontal flipping with left/right arm swapping~\cite{li2025gr}, and partial frame-skipping to synthesize speed variations demonstrated in Figure~\ref{fig:consistency}, detail in Appendix.

\section{Experiments}
\label{sec:experiments}

Our evaluation framework targets collaborative long-horizon garment manipulation, encompassing flattening from arbitrary states, folding, handover operations, and hanging.
We select this series of tasks because their contact-rich, deformable dynamics and requirement for state recovery effectively isolate and magnify the distributional shifts aforementioned.
We illustrate the detailed robot setup in Figure~\ref{fig:robot_setup}. 
We 
systematically
investigate the following research questions:
\begin{enumerate}
    \item \textbf{System Efficacy breakdown.} How do the individual components
    synergize to enhance overall performance, or do they conflict when integrated?
    (Sec.~\ref{exp:overall})

    \item \textbf{Model Arithmetic.} Can 
    MA
    of subset-trained candidates outperform both a single best candidate among them and a full-data-trained candidate? 
    Which validation split (in-domain vs. OOD) 
    demonstrates robust statistical superiority
    among different strategies?
    (Sec.~\ref{exp:arithmetic})
    \item \textbf{Stage Advantage.} Does predicting stage-conditioned advantage offer more stable supervision than value-difference baseline (RECAP in $\pi^*_{0.6}$~\cite{black2025pistar}), and how does this translate to policy success? (Sec.~\ref{exp:advantage})
    \item \textbf{Train-Deploy-Alignment.} Does expanding $P_{\text{train}}$ via Heuristic DAgger improve performance while incurring only a marginal increase in retry cost compared to standard DAgger?
    How do different control methods work with spatio-temporal data-augmentation? (Sec.~\ref{exp:consistentcy})

\end{enumerate}

\subsection{Evaluation Tasks and Metrics}
\label{sec:exp_setup}
We evaluate our approach on three challenging garment manipulation tasks with varying complexity.
\noindent\textbf{Task A: T-Shirt Flattening and Folding (Easy).} A simplified variant of the standard laundry task from the $\pi$ series~\cite{black2024pi0,black2025pi05,black2025pistar}. The robot must flatten a T-shirt from an arbitrary initial configuration and fold it. Success is defined as placing the fully folded T-shirt in the table center within 180 seconds.

\noindent\textbf{Task B: Conditional Retrieval and Sorting (Medium).} An extension of the $\pi$ series task~\cite{black2024pi0,black2025pi05,black2025pistar} involving conditional logic. The system retrieves and flattens either a T-shirt or a collared shirt from variable initial states. T-shirts must be folded and stacked (top-left), while collared shirts must be handed over to the right side, both within 180 seconds.

\noindent\textbf{Task C: Garment Hanging (Hard).} Extended from GR-3~\cite{cheang2025gr3}, this task requires retrieving the flattened collared shirt from Task B and hanging it on a rack. Success is defined as the stable suspension of the garment on the rod without dropping.

We report four metrics (mean $\pm$ accumulative standard error), calculated over 10 trials for each of three garment types.
\textbf{Success Rate (SR)} measures the percentage of trials that complete the task successfully 
(higher is better).
\textbf{Throughput (TP)} quantifies 
the estimated number of tasks completed 
per 
hour (higher is better).
\textbf{Retry Cost} is the average number of action retries per episode during evaluation (lower is better).
\textbf{Average Score} is derived from a rule-based evaluation protocol. We define subtask-specific milestones and assign partial credit upon their completion; credits are then normalized to 100.
The Appendix lists metric calculation and task details.

\begin{figure}[t]
  \centering
  \includegraphics[width=1.0\linewidth]{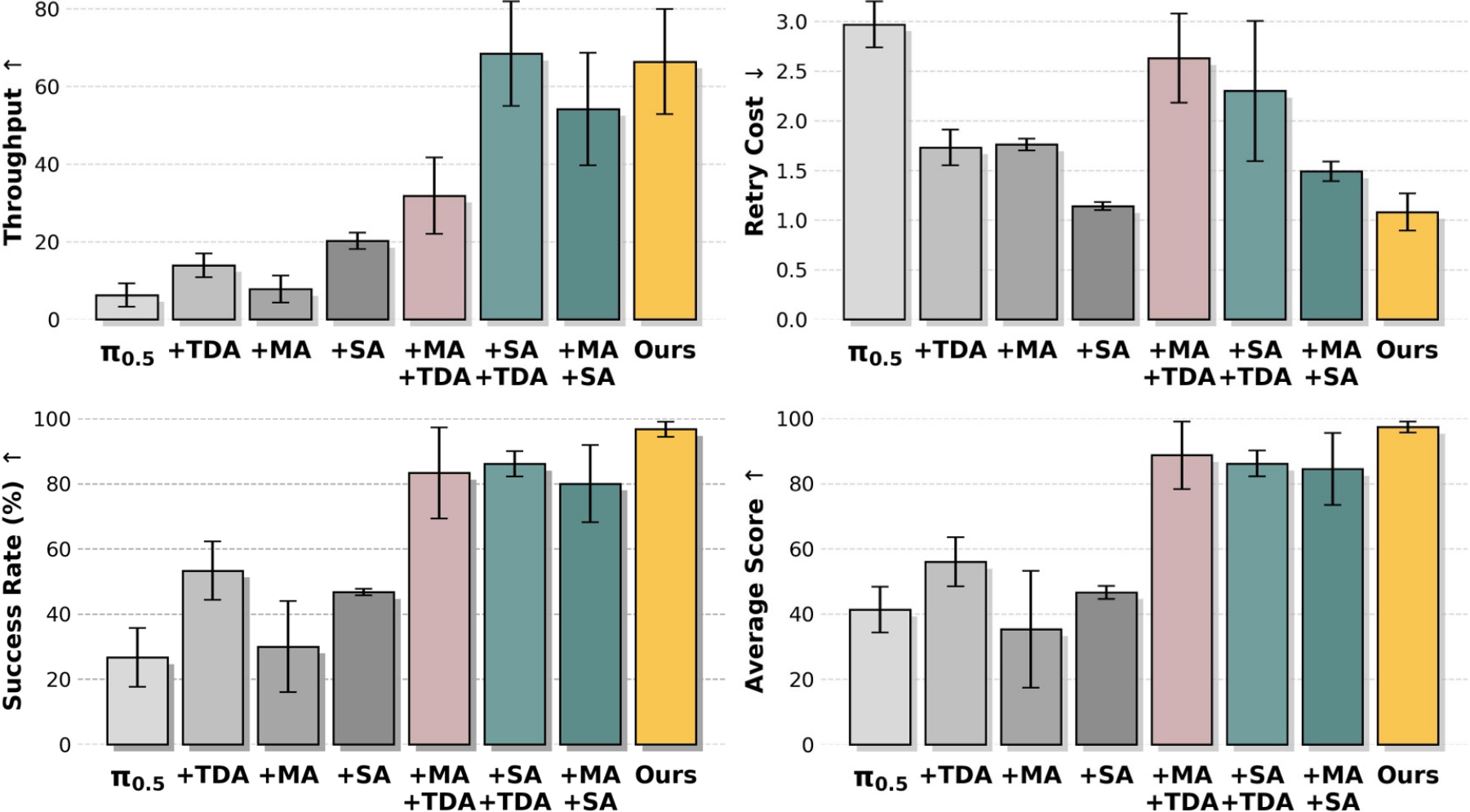}
    \caption{\textbf{\name{} system efficacy on Task A.}
    Performance improves with individual modules and increases further in pairwise settings, maximizing with the complete \name{} system (ours).
    }
  \label{fig:exp_triple}
  \vspace{-1em}
\end{figure}

\subsection{Data collection and Training Strategy}
We curate $\sim$20 hours of expert demonstrations per task, capturing diverse garment states (color, material), initial states, and environmental lighting.
We employ full-parameter fine-tuning on 8$\times$A100 GPUs via a flow-matching objective~\cite{black2024pi0}. See Appendix for hyperparameters and collection details.

\subsection{Baselines and Ablation Design}
\label{sec:exp_baselines}

\noindent\textbf{Base policy. }
We select $\pi_{0.5}$ as our primary base policy, complemented by $\pi_{0}$, 
as these are the only two open-source policies capable of viable performance on our tasks.
Despite their reported capabilities in similar domains, GO-1~\cite{bu2025agibot}, X-VLA~\cite{zheng2025xvla} and DexVLA~\cite{wen2025dexvla}
did not achieve tractable performance even after training on the full 20-hour dataset.
\noindent\textbf{MA Ablations.} 
We establish two baselines: a \textbf{single-best candidate} (selected from policies trained on individual subsets) and a \textbf{full-data candidate} (trained on the aggregated dataset).
We examine robust statistical superiority of in-domain and OOD validation split across merging techniques: 
\emph{Average weighting}~\cite{wortsman2022model} (assigning uniform weights by setting \(\alpha_i = 1/n\));
\emph{Inverse-loss}~\cite{maiti2025souper} (inversely proportional to per-checkpoint validation losses \(L_i\), setting $\alpha_i \propto 1/(L_i+\epsilon)^p$ after normalization);
\emph{Gradient descent and its adaptive invariant}~\cite{ilharco2022editing} (softmax-parameterized coefficients \(\mathbf{\alpha} = \text{softmax}(w)\) 
by minimizing the merged validation loss validation loss \(\mathcal{L}_{val}(\sum_i \alpha_i \theta_i)\) of \(\theta_{\textbf{merged}}\) via iterative updates);
and \emph{Greedy search}~\cite{wortsman2022model} (iteratively adding checkpoints that reduce validation loss most under uniform averaging among candidates).

\noindent\textbf{SA Ablations.} We compare against a self-implemented RECAP~\cite{black2025pistar} baseline implemented on $\pi_{0.5}$~\cite{black2025pi05} PaliGemma which trained to estimate progress given current frame. The advantage signal is derived from the value(progress) difference relative to a 50-step future horizon within the same trajectory.

\noindent\textbf{TDA Ablations.} 
Our temporal chunk-wise smoothing is compared against Synchronous/Asynchronous Inference~\cite{tang2025vlash}, temporal ensembling~\cite{zhao2023aloha}, and RTC~\cite{black2025rtc}. 
We evaluate control methods across different spatio-temporal augmentation settings to identify the optimal configuration.
Additionally, we assess the impact of data augmentation by comparing standard DAgger~\cite{ross2011dagger} with our Heuristic DAgger, examining both the performance delta and the generalization of these methods across $\pi_{0}$~\cite{black2024pi0} and $\pi_{0.5}$~\cite{black2025pi05} architectures.

\begin{figure}[t!]
  \centering
  \includegraphics[width=1.0\linewidth]{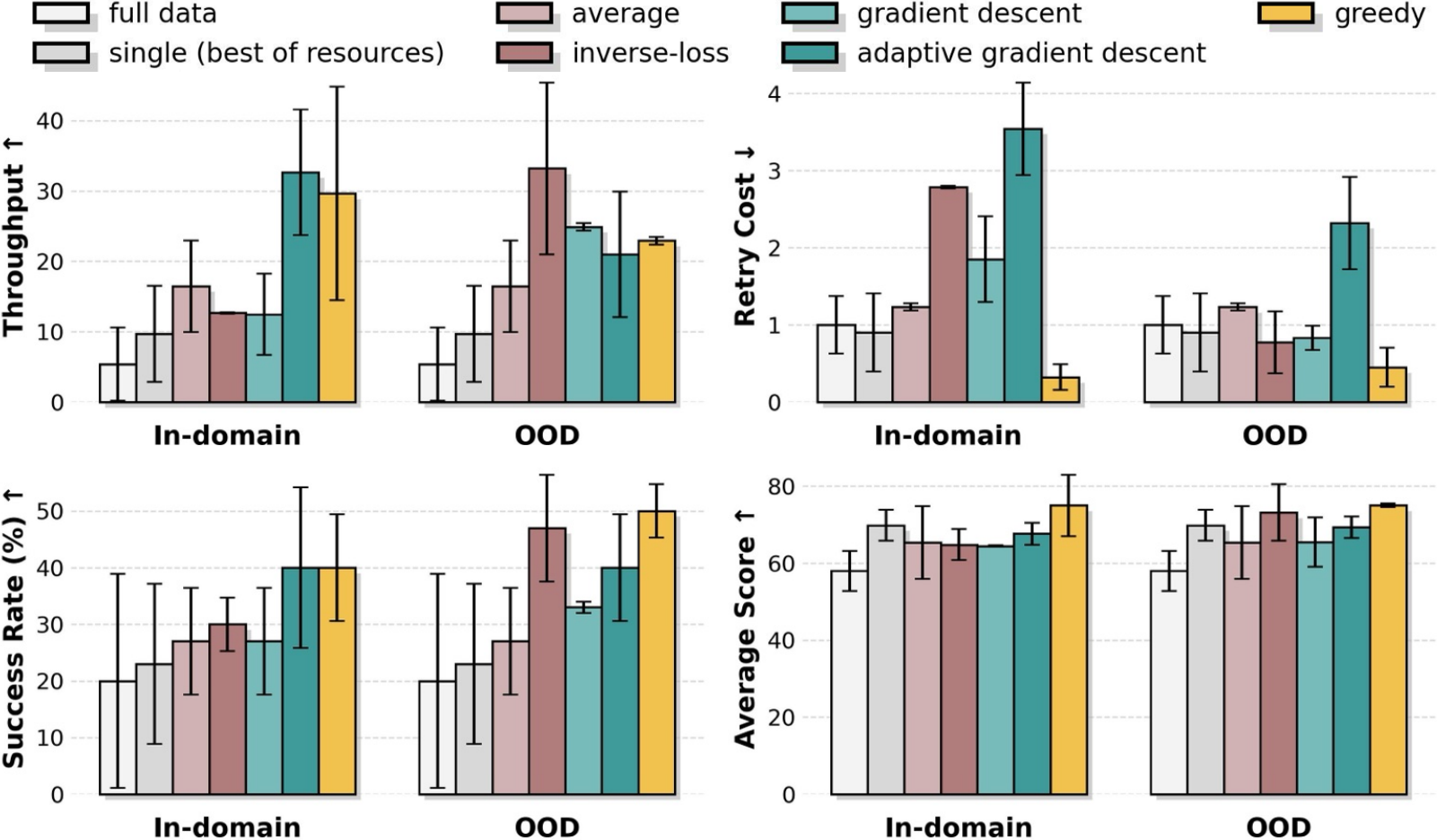}
    \caption{
    \textbf{Ablations of MA on Task C.} All MA variants outperform single-best and full-data candidates in throughput and success rate with robust statistical significance, despite increased retry costs in some implementations. Furthermore, OOD validation demonstrates enhanced stability and reduced standard error relative to in-domain validation.
    }

  \label{fig:exp_souping}
  \vspace{-3pt}
\end{figure}

\begin{figure}[t!]
  \centering
  \includegraphics[width=1.0\linewidth]{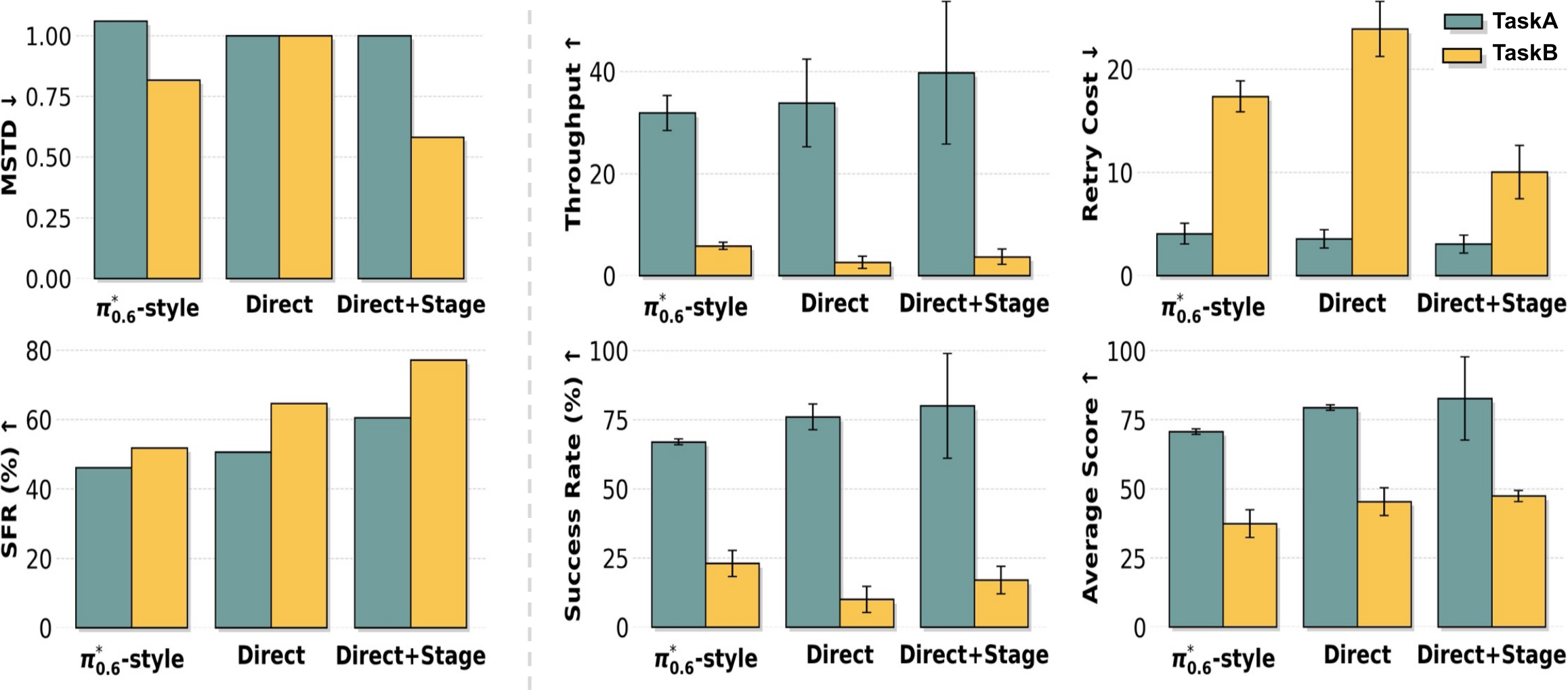}
    \caption{\textbf{Ablations of SA on Task A \& B.} 
    Left: quantifying
    numerical stability via Smooth Frame Ratio (SFR) and Mean Squared Temporal Difference (MSTD).
    Right: SA shows superior stability against the $\pi^*_{0.6}$-style advantage baseline, which correlates positively with the observed performance gains.
    }
  \label{fig:exp_advantage_ff}
  \vspace{-1em}
\end{figure}

\subsection{\name{} System Efficacy Breakdown}
\label{exp:overall}

In Figure~\ref{fig:exp_triple}, we report the performance breakdown of each module in the \name{} system on task A. 
By selecting the optimal configuration for each module (MA, SA, TDA)—as detailed in subsequent sections—we ensure effective system integration, where performance scales monotonically with components added.
Specifically, SA is the dominant factor for throughput, whereas TDA drives the success rate but incurs higher retry costs. This aligns with our insight that TDA encourages persistent retrying—a behavior that naturally improves task completion at the expense of increased operational cost.

\subsection{Model Arithmetic Results}
\label{exp:arithmetic}

Figure~\ref{fig:exp_souping} presents a comprehensive analysis across all metrics, benchmarking MA variants against non-MA baselines and quantifying the performance differentials between distinct MA strategies.
\textbf{1)} First, all MA variants outperform both the single-best candidate and the full-data baseline, validating the efficacy of the approach. Notably, merging weights from subset-trained models surpasses training on the combined dataset (joint training). This suggests that fine-tuned VLAs may exhibit \textit{extreme parameter redundancy}, akin to phenomena observed in LLMs~\cite{yu2024language}.
\textbf{2)} Second, OOD validation data proves to be a more robust selection criterion than in-domain data, yielding lower standard errors and higher performance across all metrics. 
This finding supports our hypothesis that OOD data (e.g., DAgger) effectively bridges the gap between $P_\text{train}$ and $P_\text{test}$. 
Consequently, utilizing DAgger data to calibrate mixing weights ensures that $Q_\text{model}$ prioritizes modes in $P_\text{train}$ that align with deployment dynamics.
Among MA strategies, greedy search proves most effective across diverse settings.
This reinforces our finding that validation loss on DAgger data accurately reflects distributional gaps, enabling $Q_\text{model}$ to improve mode coverage for test-time generalization~\cite{chen2025retaining}.

\begin{figure}[t]
  \centering
  \includegraphics[width=1.0\linewidth]{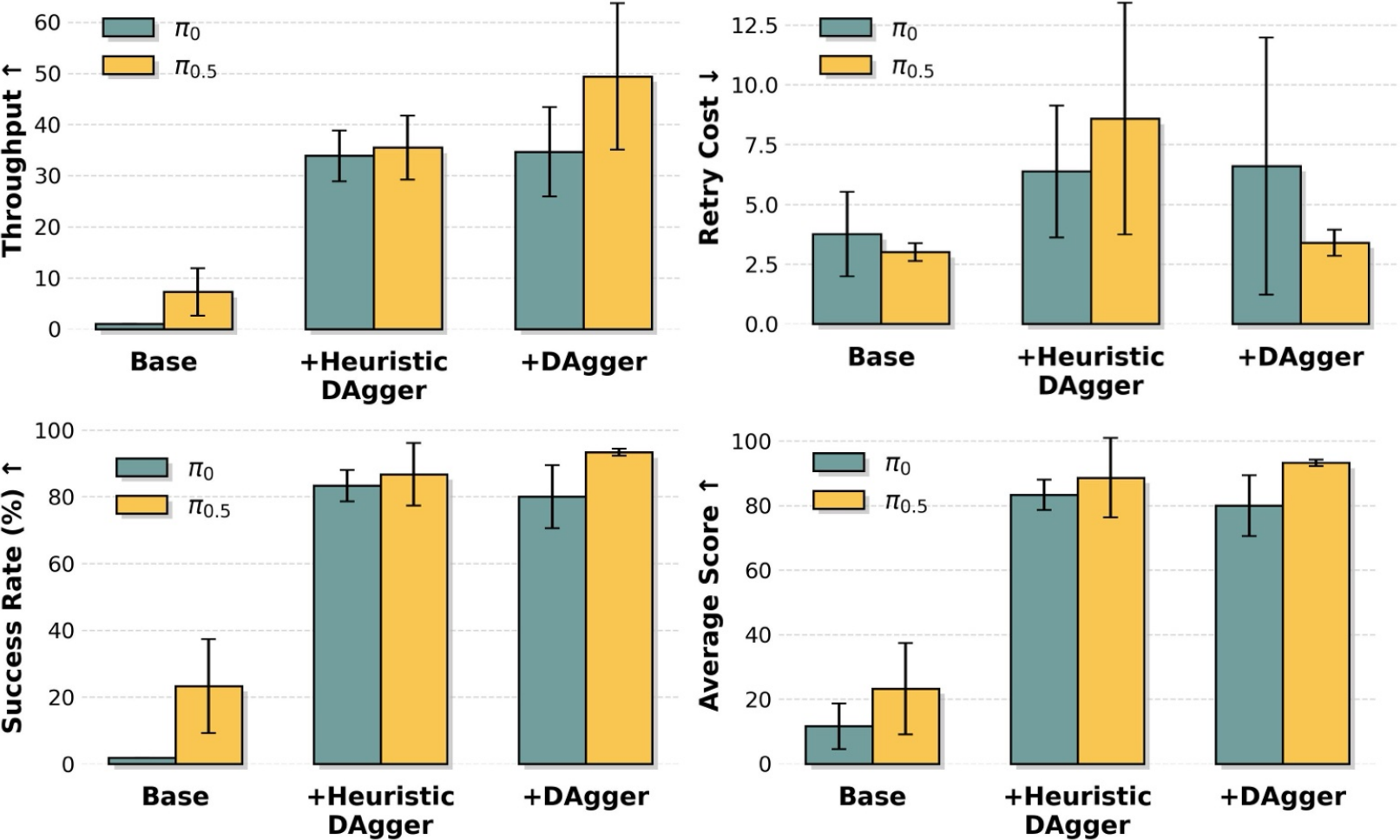}
    \caption{\textbf{Ablations of Heuristic and standard DAgger on Task A.} All DAgger-style variants improve throughput, success rate, and score, despite higher retry costs under heuristic DAgger. For $\pi_{0.5}$, full DAgger further boosts throughput while reducing retry cost, yielding a better trade-off.}
  \label{fig:exp_consistency_dagger_ff}
\end{figure}

\begin{figure}[t!]
  \centering
  \includegraphics[width=1.0\linewidth]{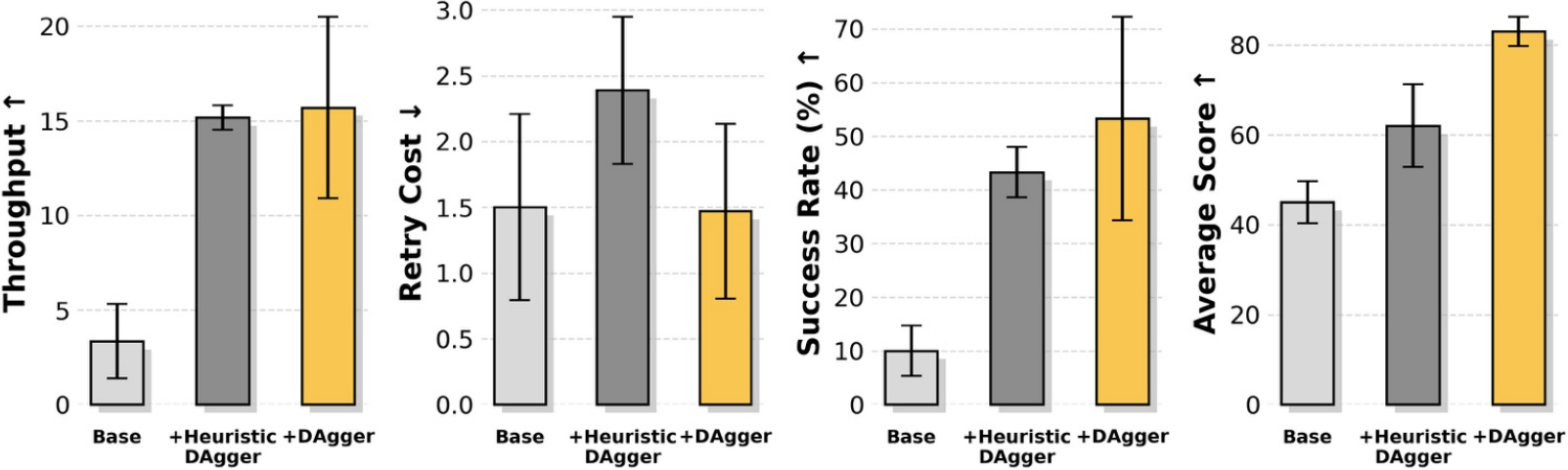}
    \caption{\textbf{Ablations of Heuristic and standard DAgger on Task C.} DAgger works in a similar trend as Figure.~\ref{fig:exp_consistency_dagger_ff}, }
  \label{fig:exp_consistency_dagger_demoB}
  \vspace{-1em}
\end{figure}

\subsection{Stage Advantage Results}
\label{exp:advantage}

Figure~\ref{fig:exp_advantage_ff} demonstrates that improved numerical stability from SA translates to broad performance gains.
Specifically, SA minimizes retry overhead in Task B (long-horizon, conditional). This confirms that SA effectively biases the policy against idling and spurious retries during deployment ($P_\text{test}$), ensuring consistent task progression. 
The results also underscore that direct advantage estimation requires stage-aware labels to ensure robustness.

\begin{figure}[t!]
  \centering
  \includegraphics[width=1.0\linewidth]{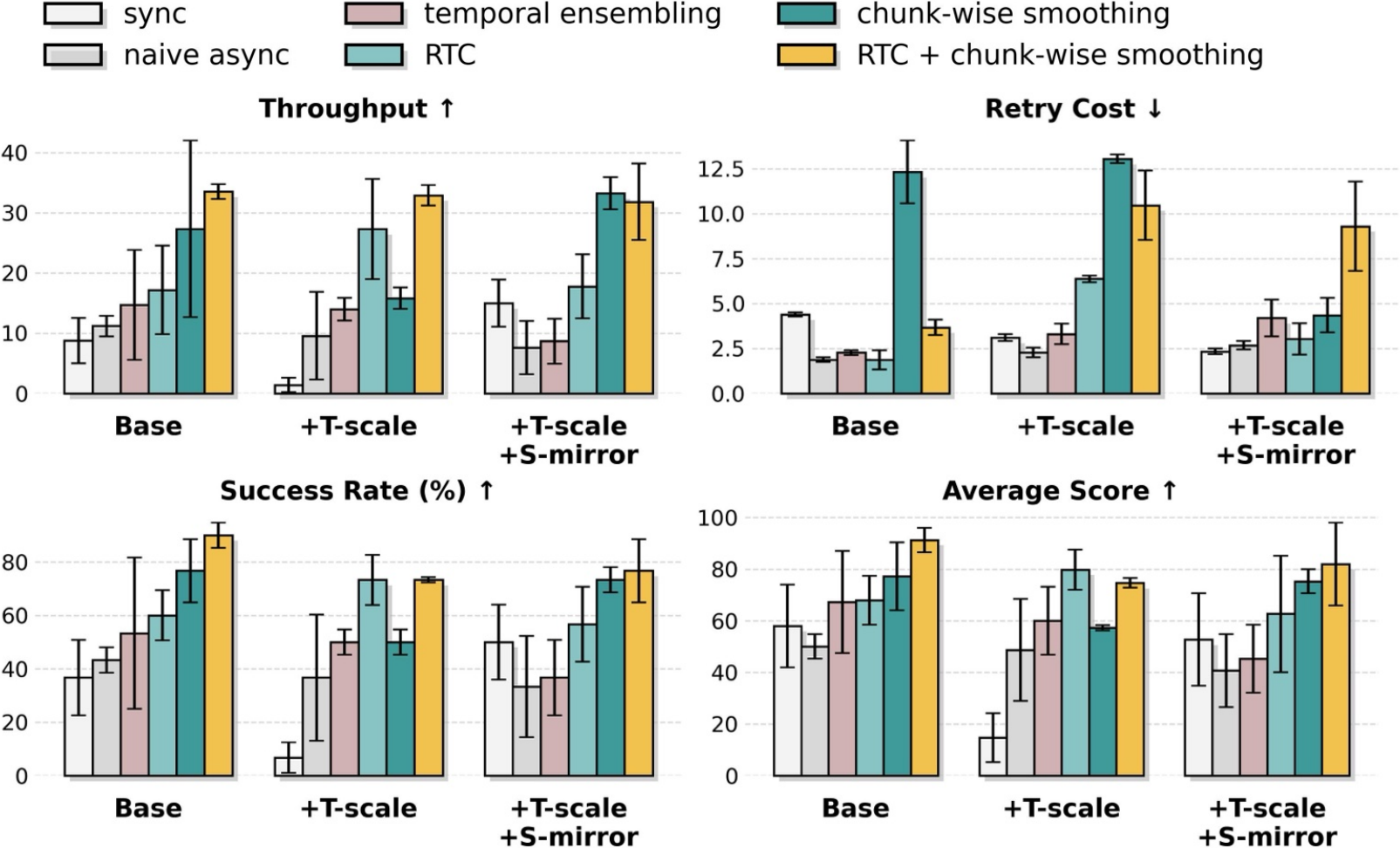}
    \caption{
    \textbf{Ablations of control strategies and spatio-temporal augmentation on Task A.}
    Temporal chunk-wise smoothing outperforms temporal ensembling, RTC; Combining our method with RTC further improves the performance.
    }
  \label{fig:exp_consistency_aug&control}
  \vspace{-1em}
\end{figure}

\subsection{Train-Deploy-Alignment Results}
\label{exp:consistentcy}

Figure~\ref{fig:exp_consistency_dagger_ff} and \ref{fig:exp_consistency_dagger_demoB} evaluate Heuristic and standard DAgger across models ($\pi_{0.5}$, $\pi_0$) and tasks (Task A, Task C).
\textbf{1)} Heuristic DAgger significantly improves failure recovery and overall performance, achieving substantially higher SR/TP compared to the baseline. Note that base's low recovery cost reflects absent self-correction behaviors rather than efficient recovery.
\textbf{2)} Heuristic DAgger provides comparable recovery quality to DAgger, while inference-free makes it more cost-efficient to address failure cascade.
Fig.~\ref{fig:exp_consistency_aug&control} benchmarks temporal chunk-wise smoothing against other control strategies across different augmentation settings. 
\textbf{1)} Temporal chunk-wise smoothing outperforms temporal ensembling and RTC in most cases,
demonstrating its effectiveness in translating \(Q_{\text{model}}\) capability into \(P_{\text{test}}\) performance.
\textbf{2)} Spatio-temporal augmentation shows task-dependent effect, providing no significant gain on Task A.

\section{Conclusion and Limitations}
\label{sec:conclusion}
We present \name{} to address distributional shifts across the robot learning.
Through Model Arithmetic, Stage Advantage, and Train-Deploy-Alignment, the system targets key failure modes including coverage deficiency and temporal mismatch,
which is validated extensively on complex garment manipulation tasks 
and yields robust long-horizon performance.

Several limitations are spot through this journey of achieving 100\% reliability in robotic manipulation. The first is 
\textbf{Scalability.} 
The prominence of robot foundation models hinges on their promise of broad generalization.
This study, however, did not explicitly evaluate the retention of pre-trained priors during the post-training process.
Future research should address the retention of pre-trained priors and evaluate the extensibility of \name{} to rigid-body manipulation tasks.
Furthermore, it remains to be seen if Model Arithmetic can integrate distinct task policies—rather than merely subsets—to advance general-purpose robotics.
The second is
\textbf{Data Valuation.} Our results confirm that data utility is highly variable. Currently, validating data quality requires costly full-training loops or slow, non-parallelizable replay checks (where successful replay indicates high utility). Establishing efficient, predictive metrics for data quality without incurring these overheads remains a key challenge for future research.

\section*{Acknowledgements}

We thank our robot operators for data collection, evaluations, logistics, and video recording, and our technicians for robot maintenance and repair.
We thank Longyan Wu for refining teaser and robot setup figure.
We thank Zhiqian Lan, Jiaheng Wang for their early help in this project, including data generation pipeline construction as well as inference smoothness optimization.
We thank Jiazhi Yang, Yixuan Pan, Yinghui Li, Junli Ren, Haoran Jiang, Kunyang Lin, Wencong Zhang, Jinwei Li, Kai Zhang for discussion and feedback.

\bibliographystyle{plainnat}
\bibliography{references}

\clearpage
\newpage
\onecolumn
\appendix

This supplementary material provides detailed analysis to substantiate our main findings. Section~\ref{sec:supp-questions} poses motivating questions to offer alternative perspectives on the work and might address potential questions. 
Section~\ref{sec:supp_related} reviews broader related work on robustness from RL in manipulation. 
Section~\ref{sec:supp_method} elaborates on system implementation and design alternatives, accompanied by the code implementation in the related code files. 
Section~\ref{sec:supp_exp_setup} details experimental protocols—including hyperparameters, hardware configurations, and training strategies—with extended results provided in Section~\ref{sec:supp_exp_tables}.
Finally, Section~\ref{sec:supp_failure} analyzes failure modes, Section~\ref{sec:supp_data_ethics} outlines data ethics and corresponding licensing, and Section~\ref{sec:supp_contributions} lists the contribution.

\subsection{Motivating Questions}
\label{sec:supp-questions}

These are the questions that might be raised by the audience and the ones we think beyond this project:

\bigskip
\noindent\textbf{Q1.} \textit{What is the relationship between stage advantage and RL?}
\smallskip

Our stage advantage follows the paradigm of advantage-weighted regression~\cite{peng2019advantage}, which could be seen as an offline RL~\cite{sutton1998reinforcement} where the monte carlo reward is calculated directly from data.
However, as observed in the huge variance in the learned reward, advantage-weight diffusion policy~\cite{frans2025diffusion} and its massive-scale validation on $\pi^*_{0.6}$~\cite{black2025pistar} parameterize a Q network instead from the demonstration data.
Stage advantage follows the paradigm of $\pi^*_{0.6}$, while improve from the baseline with more stable numerical result in advantage estimation and thus results in better performance of the trained policy.

\bigskip
\noindent\textbf{Q1.} \textit{Why not online RL such as PPO?}
\smallskip

PPO~\cite{schulman2017PPO} and its adaptation on diffusion-based policy such as DSRL~\cite{wagenmaker2025dsrl} stand as another trend in RL to directly learn from the interaction with the environment, which is referred as online RL.
These methods are fundamentally limited by real-world sample inefficiency. Unlike simulation, scaling physical experiments is constrained by the prohibitive costs of parallelization and resetting, constituting a major bottleneck.
In contrast, AWR~\cite{peng2019advantage} incurs only a marginal labeling cost to bias the learning objective toward high-progress actions, thereby maximizing the utility of both human demonstrations and policy rollouts.

\bigskip
\noindent\textbf{Q2.} \textit{Why choosing $\pi_{0.5}$ as primary baseline? }
\smallskip

To identify a robust starting point, we conducted preliminary evaluations of several prominent VLA models, including X-VLA~\cite{zheng2025xvla}, GO-1~\cite{bu2025agibot}, UniVLA~\cite{bu2025univla}, and OpenVLA~\cite{kim2024openvla}. Despite claims of broad adaptability and robustness, these models failed to generalize to our experimental tasks.
Even after extensive post-training with aligned hardware settings and exhaustive hyperparameter tuning---where training loss converged satisfactorily---the policies achieved negligible success rates across 30 distinct trials (10 trials per garment).
We provide these failure videos in this supplementary as well.
Consequently, we excluded these models from the primary comparison to ensure our analysis focused on baselines capable of meaningful task completion.

\bigskip
\noindent\textbf{Q4.} \textit{How to define/design indicative/informative advantage?}
\smallskip

This is a critical direction for future work. Naive behavior cloning treats all demonstrated actions equally, despite many actions lacking explicit utility.
Training on these non-informative segments can bias the policy toward meaningless behaviors, resulting in local optima where the robot repeats actions without making progress.
Advantage weighting offers a mechanism to filter these behaviors by emphasizing actions that yield high returns. However, our current formulation relies on a heuristic proxy---using temporal progress to label advantage---which assumes task completion is strictly monotonic. 
A more robust solution would replace this heuristic with a unsupervised advantage estimator capable of distinguishing truly instrumental actions from noise, independent of temporal linearity.

\bigskip
\noindent\textbf{Q5.} \textit{How to define good robot data?}
\smallskip

Empirically, we find that data quality is a governing factor in policy performance, accounting for success rate fluctuations between 20\% and 60\% under identical settings.
Notably, we observe a ``masking effect" where certain algorithmic improvements yield significant gains on low-quality data but saturate or vanish when trained on high-quality demonstrations. 
Data quality seems to be the core part under current trend of imitation learning, yet not many literature seriously discuss that.
To address this overlooked variable, we state replay-ability as one of the most important principles for data validity.
A trajectory is replay-able if open-loop re-execution leads to task completion (or significant progress) from a similar initial state.
This criterion not only filters corrupted demonstrations but also serves as a diagnostic tool, exposing systematic hardware inconsistencies between the data collection $P_{\text{train}}$ and inference $P_\text{test}$ environments.

\bigskip
\noindent\textbf{Q6.} \textit{What is the common failure case?}
\smallskip

\begin{figure}[th!]
  \centering
  \includegraphics[width=\columnwidth]{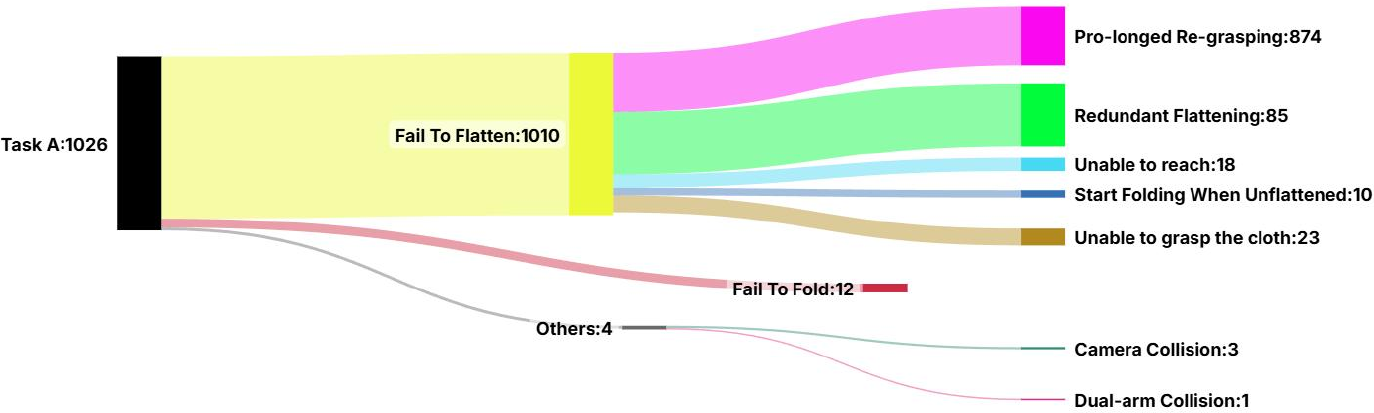}
  \vspace{-0.5em}
  \caption{\textbf{Visualization of Failure Case in Task A.}}
  \label{fig:supp_failure_case}
  \vspace{-1em}
\end{figure}

We observe two primary failure modes: 
\textbf{1) Spatial misalignment}, where the policy fails to localize the correct grasping affordance; and 
\textbf{2) Policy stagnation}, where the robot enters a "dead-loop" of repetitive, non-productive actions.
Both are illustrated in Figure~\ref{fig:supp_failure_case}.
These failures underscore two fundamental deficiencies in current pre-trained models: a lack of fine-grained spatial understanding and insufficient long-horizon task planning.
While we partially mitigate the former via Model Soups and the latter through Stage Advantage, these are extrinsic corrections. 
We argue that future pretrained weights must possess stronger intrinsic capabilities for spatial grounding and sequential logic to fully resolve these issues.

\bigskip
\noindent\textbf{Q6.} \textit{What is the ground difference between different ``robotic foundation models''?}
\smallskip

Resonating with our discussion on failure modes, the core distinction lies not in the models' zero-shot capabilities, but in their fine-tuning dynamics---specifically, their plasticity in acquiring novel spatial understanding and task planning skills during post-training. 
We observe that certain architectures (e.g., $\pi_0$ and $\pi_{0.5}$ demonstrate significantly higher adaptability than others, suggesting a more robust initialization from pre-train weight for downstream learning.
Consequently, the field requires new metrics that benchmark the intrinsic representational quality of these foundation models, rather than relying solely on success rates in simplistic evaluation environments.

\subsection{More Related Work}
\label{sec:supp_related}

\subsubsection{Discussion about RL for robustness}

Reinforcement learning has become a prevalent approach for post-training VLA foundation policies to improve manipulation robustness and precision, with applications spanning both simulation \cite{liu2023libero, mu2025robotwin, mees2022calvin, lu2025vla, li2025simplevla, chen2025tgrpo} and the real world \cite{luo2025hil-serl, xiao2025PLD, luo2024serl}.
A central challenge lies in fine-tuning large pre-trained models without destabilizing the learned representations.
Advantage-conditioned approaches \cite{black2025pistar, frans2025diffusion, kumar2019reward} allow full-model optimization by conditioning generation on estimated advantages, circumventing the need to differentiate through diffusion or flow-matching denoising processes \cite{lei2025rl}.
The effectiveness of such methods hinges on the quality of the advantage signal.
A popular recipe employs vision-language models to score task progress \cite{ma2024gvl, zhang2025rewind, zhai2025vlac, ghasemipour2025self}, from which advantages are recovered as $A(s,a)=V(s')-V(s)$ \cite{black2025pistar} and used in advantage-weighted regression \cite{wu2023elastic, kuba2023advantage}.
This difference-of-values formulation, however, compounds independent estimation errors into noisy training signals, and the lack of stage awareness causes the value function to produce ambiguous predictions when visually similar states appear at different phases of a multi-stage task \cite{black2025pistar, li2025gr}.
Our method addresses both limitations: instead of subtracting two noisy value predictions, we directly model advantage as a single pairwise prediction $f_\theta(s, s')$, yielding lower-variance signals by construction; and instead of estimating progress under a single global objective, our Stage Advantage conditions on the current stage goal, explicitly resolving multi-valued ambiguity and providing accurate, stage-aware supervision for long-horizon policy optimization.

\subsubsection{Control method discussion}
While asynchronous inference reduces latency, existing methods face significant implementation hurdles. SmolVLA~\cite{shukor2025smolvla} employs a naive chunk-switching strategy, resulting in severe prediction-execution misalignment and control instability. 
Similarly, the concurrent A2C2~\cite{sendai2025leave} addresses misalignment by adding auxiliary correction heads, which necessitates architectural modifications. 
In contrast, our method augments asynchronous inference with temporal chunk-wise smoothing, incurring negligible computational or architectural overhead.

\subsection{Method Details}
\label{sec:supp_method}
\subsubsection{why we call it $\chi_0$}
Our approach is grounded in the principle of \textbf{K}inetics \textbf{A}ligned to \textbf{I}ntelligence. Here, ``Kinetics'' defines the operational domain—spanning training ($P_{\text{train}}$) and deployment ($P_{\text{test}}$)—while ``Intelligence'' denotes the inductive bias encoded in the parameter space ($Q_{\text{model}}$). We designate this initial iteration as KAI\_0. Noting the phonetic resemblance between ``KAI'' and the Greek letter $\chi$, we name the system \name{}, paying respect to the state-of-the-art $\pi$ policy series.

\subsubsection{MA details}

We combine 4 checkpoints via Model Arithmetic, each trained on a different subset of the training data.
Checkpoints are weighted using inverse loss weighting: validation loss is computed as the mean prediction error on a held-out set, and checkpoints with lower loss receive proportionally higher weights.
We additionally consider two baselines: a \emph{single best} candidate, selected as the individual checkpoint with the lowest validation loss, and a \emph{full data} candidate, trained on the complete merged dataset across all subsets.

\subsubsection{SA details}

The advantage estimator is trained on pairs of frames sampled from the same episode at arbitrary timestamps, with the supervision target being their relative progress, defined as the subtraction of two states over their timestamps in an episode.
When training \name{}, we set the threshold $\epsilon = 0.3$ following the hyperparameter in $\pi^*_{0.6}$. Training samples are ranked by their advantage values, and the top $\epsilon$ fraction is labeled as positive while the rest is treated as negative.
Stages correspond to semantic sub-goals within each long-horizon task: two stages for task A (flattening, folding), four for task B (retrieving, flattening, folding, handover), and three for task C (retrieving, dressing the rack, hanging).

\subsubsection{TDA details}

Standard DAgger data collection proceeds as follows:
\textbf{(1)} Initialize the environment and start policy inference.
\textbf{(2)} When the policy fails repeatedly---defined as failing to grasp the clothes for more than 10 consecutive attempts, or producing repetitive action patterns without progress for more than 10 steps---it is interrupted. On the software side, the policy stops receiving new observations and produces no further actions; on the hardware side, the robot is held at its current state and switched to tele-operation mode to accept human correction.
\textbf{(3)} Both the human correction and the preceding policy rollout are saved.
Heuristic DAgger simplifies this process by treating the failure state at the point of interruption as the initial state for collecting human expert demonstrations, without requiring online policy execution.

\subsection{Experimental Details}
\label{sec:supp_exp_setup}

\begin{table*}[t]
\centering
\begin{minipage}[t]{0.48\textwidth}
    \centering
    \caption{Hyper-parameters for fine-tuning.}
    \label{tab:supp_hyperparams}
    \begin{tabular}{l|c}
    \toprule
    \textbf{Hyperparameter} & \textbf{Value} \\
    \midrule
    \multicolumn{2}{l}{\textit{Data \& Input}} \\
    Expert demonstration hours & $\sim$20 h per task \\
    Action chunk length $K$ & 50 \\
    Execution frequency & 100 Hz \\
    \midrule
    \multicolumn{2}{l}{\textit{Optimization}} \\
    Training steps & 80{,}000 \\
    Batch size & 128 \\
    Optimizer & AdamW \\
    Learning rate & 2.5$\times$10$^{-5}$ \\
    Cosine Decay Steps & 10{,}000 \\
    Conditioned noise level $\sigma$ & [0.001, 1.0] \\
    Gradient Clip & 1.0 \\
    \midrule
    \multicolumn{2}{l}{\textit{Module-Specific}} \\
    MA: Number of checkpoints & 4 \\
    SA: Advantage threshold $\epsilon$ & 0.3 \\
    \midrule
    \multicolumn{2}{l}{\textit{Infrastructure}} \\
    Training GPUs & 8 $\times$ A100 \\
    Inference GPUs &  RTX 4090 \\
    
    \bottomrule
    \end{tabular}
\end{minipage}
\hfill
\begin{minipage}[t]{0.48\textwidth}
    \centering
    \caption{Score standard (normalized to 100).}
    \label{tab:supp_score}
    \begin{tabular}{l|l|c}
    \toprule
    \textbf{Task} & \textbf{Sub-goals} & \textbf{Score} \\
    \midrule
    \multirow{4}{*}{\shortstack[l]{Task A\\(Easy)}}
    & Flatten garment              & +40 \\
    & 1st fold                     & +20 \\
    & 2nd fold                     & +20 \\
    & 3rd fold                     & +20 \\
    \midrule
    \multirow{5}{*}{\shortstack[l]{Task B -- T-shirt\\(Medium)}}
    & Retrieve \& flatten          & +40 \\
    & 1st fold                     & +15 \\
    & 2nd fold                     & +15 \\
    & 3rd fold                     & +15 \\
    & Stack to top-left            & +15 \\
    \midrule
    \multirow{3}{*}{\shortstack[l]{Task B -- Shirt\\(Medium)}}
    & Retrieve from basket         & +30 \\
    & Flatten                      & +50 \\
    & Pull to right-side table     & +20 \\
    \midrule
    \multirow{6}{*}{\shortstack[l]{Task C\\(Hard)}}
    & Pull garment rightward       & +15 \\
    & Grasp collar                 & +15 \\
    & Grasp hanger                 & +15 \\
    & Insert hanger into sleeve    & +20 \\
    & Hook left collar on hanger   & +20 \\
    & Hang on standing rack        & +15 \\
    \bottomrule
    \end{tabular}
\end{minipage}
\end{table*}

\subsubsection{Data, Training Strategy and Evaluation Criteria}

\begin{wrapfigure}{lb}{0.5\columnwidth} 
    \vspace{-0.8em} 
    \centering 
    \includegraphics[width=\linewidth]{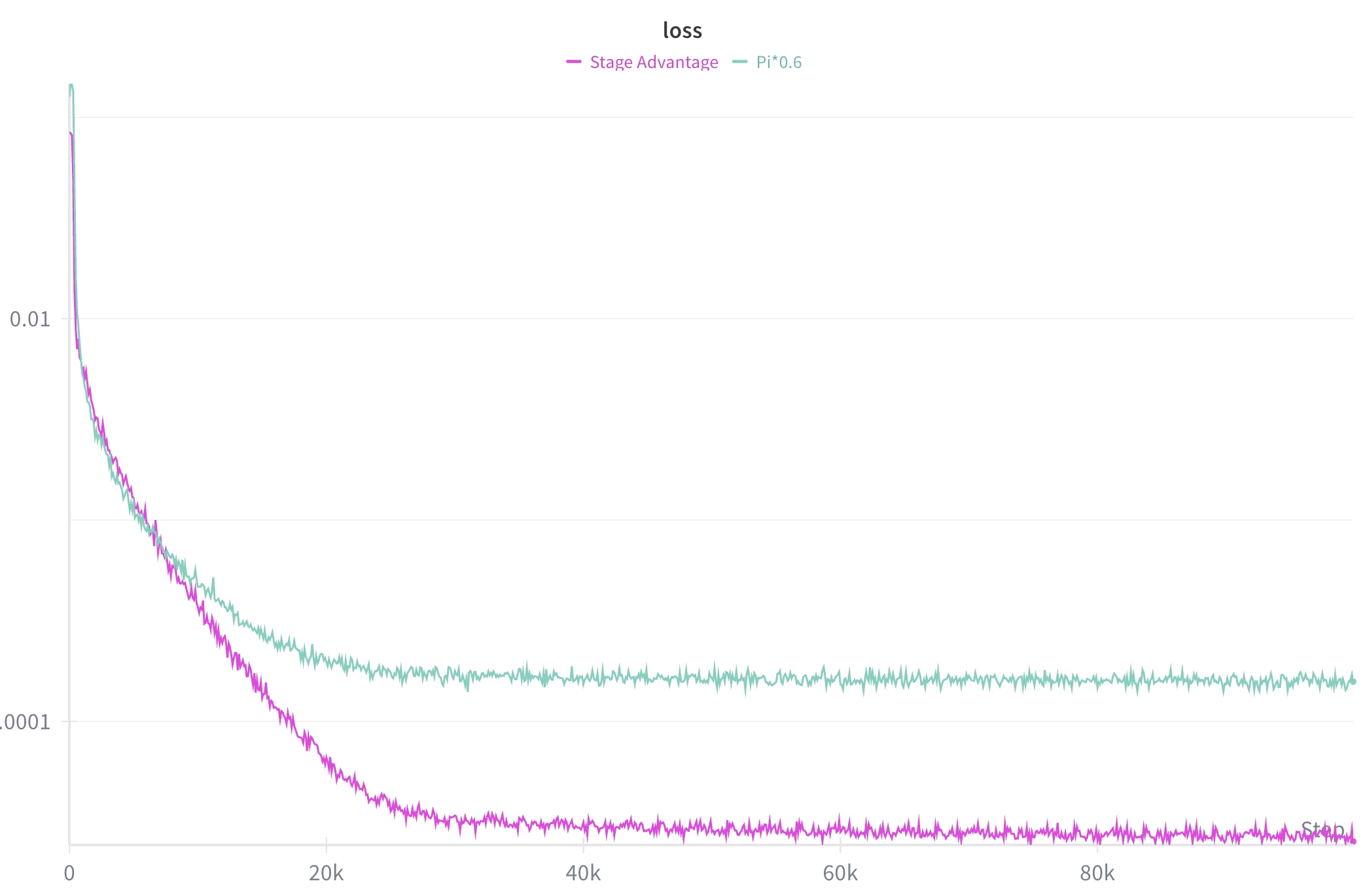} 
    \caption{\textbf{Training loss curves of SA and $\pi^*_{0.6}$-style implementation.} SA has better convergence performance than $\pi^*_{0.6}$-style implementation in terms of loss value.} 
    \label{fig:supp_training_loss_SA} 
    \vspace{-0.8em} 
\end{wrapfigure}

\begin{itemize}

    \item \textbf{Data Scale and Diversity.} Data was collected at 30 Hz under a Standard Operating Protocol (SOP) to ensure consistent execution quality and duration. The dataset comprises {2668} episodes for Task A (avg. {30.41} s), {3519} for Task B (avg. {34.40} s), and {2988} for Task C (avg. {39.01} s). Note that these means include shorter DAgger interventions. To ensure diversity, we randomize initial conditions including object state (position, crumpling configuration, size, color) and environmental factors (illumination intensity and direction). 
    \item \textbf{Training Strategy and Hyperparameters.} We conduct full-parameter fine-tuning on the open-source $\pi_{0.5}$ model using a Flow Matching objective. Each task is fine-tuned independently. Refer to Table~\ref{tab:supp_hyperparams} for a detailed list of hyperparameters.

    \item \textbf{Evaluation Criteria (Average Score).} We employ a rule-based evaluation protocol that assigns partial credit for sub-goal completion. The specific sub-goals and their corresponding scoring weights are detailed in Table~\ref{tab:supp_score}.

\end{itemize}

\subsubsection{Hardware configuration and Inference details}

As illustrated in Figure~{\textcolor{red}{6}} in the main paper, our experimental setup includes two bimanual systems: one composed of AgilexRobotics Piper arms and the other of ARXRoboticsX X5 arms. Both dual-arm platforms feature two 6-DoF arms equipped with 1-DoF parallel gripper. A key kinematic distinction lies in the 5th joint configuration: the Agilex Piper arm with a pitch joint, whereas ARX X5 utilizes a yaw joint.
For perception, each system is equipped with three Intel RealSense D435i cameras (one fixed head-view and two wrist-view) capturing $640 \times 480$ RGB images. The visual sampling rate is synchronized at $30$Hz for both data collection and inference, while the low-level joint controllers operate at a higher frequency of $100-200$ Hz. All inference are performed on a Ubuntu 20.04 workstation equipped with an NVIDIA RTX 4090 GPU.

\subsubsection{Empirical loss curve}

Figure~\ref{fig:supp_training_loss_SA} compares the training loss curves. 
The proposed Stage Advantage (SA) demonstrates superior convergence characteristics relative to the $\pi^*_{0.6}$ baseline. 
This empirical evidence suggests that SA offers enhanced numerical stability within the advantage-weighted behavioral cloning framework.

\begin{figure}[t]
  \centering
  \subfloat[\textbf{Ablations of MA on Task A}\label{fig:exp_soup_ff}]{
    \includegraphics[width=0.48\columnwidth]{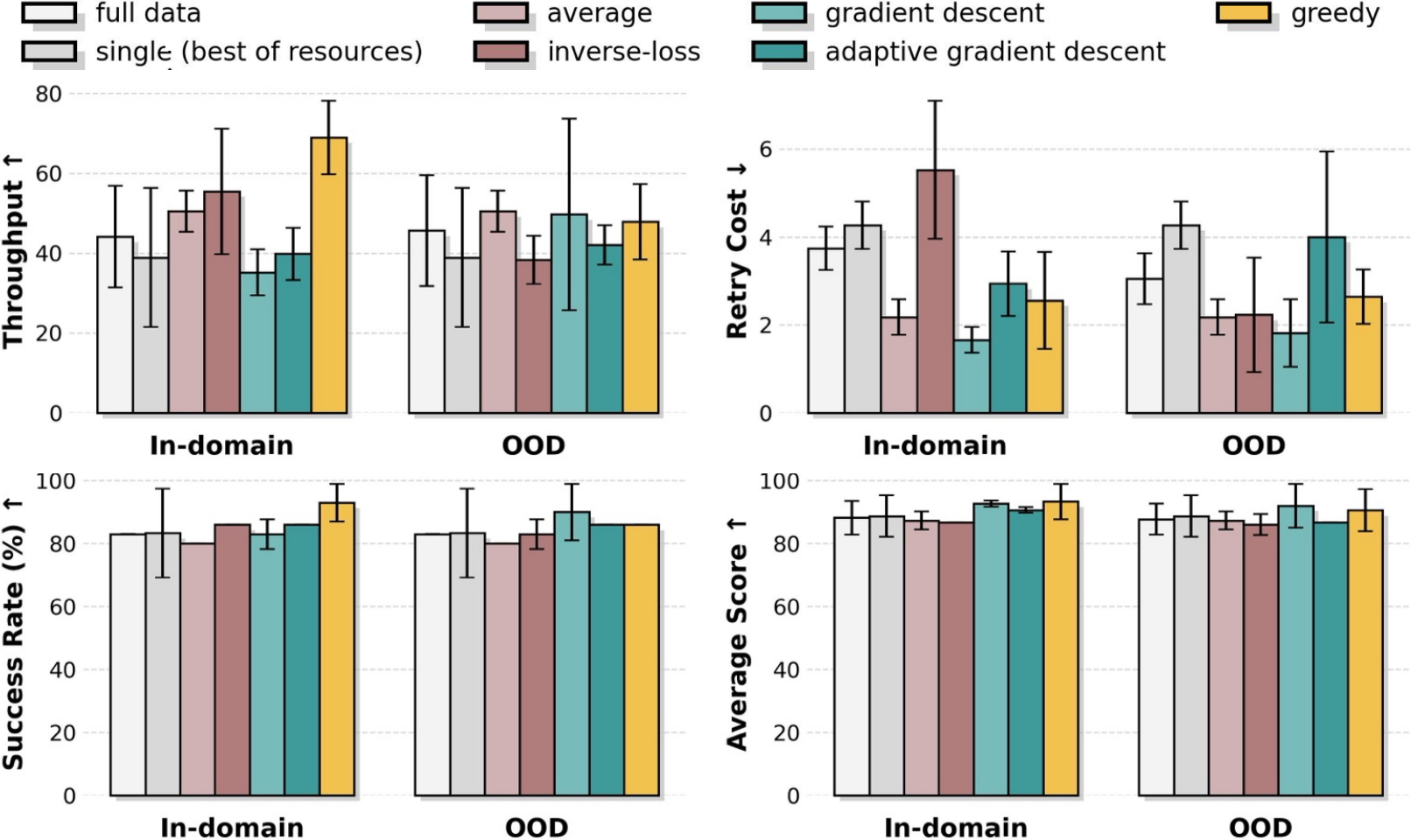}
  }
  \hfill
  \subfloat[\textbf{Ablations of MA on Task B}
    \label{fig:supp_exp_soup_hang}]{
    \includegraphics[width=0.48\columnwidth]{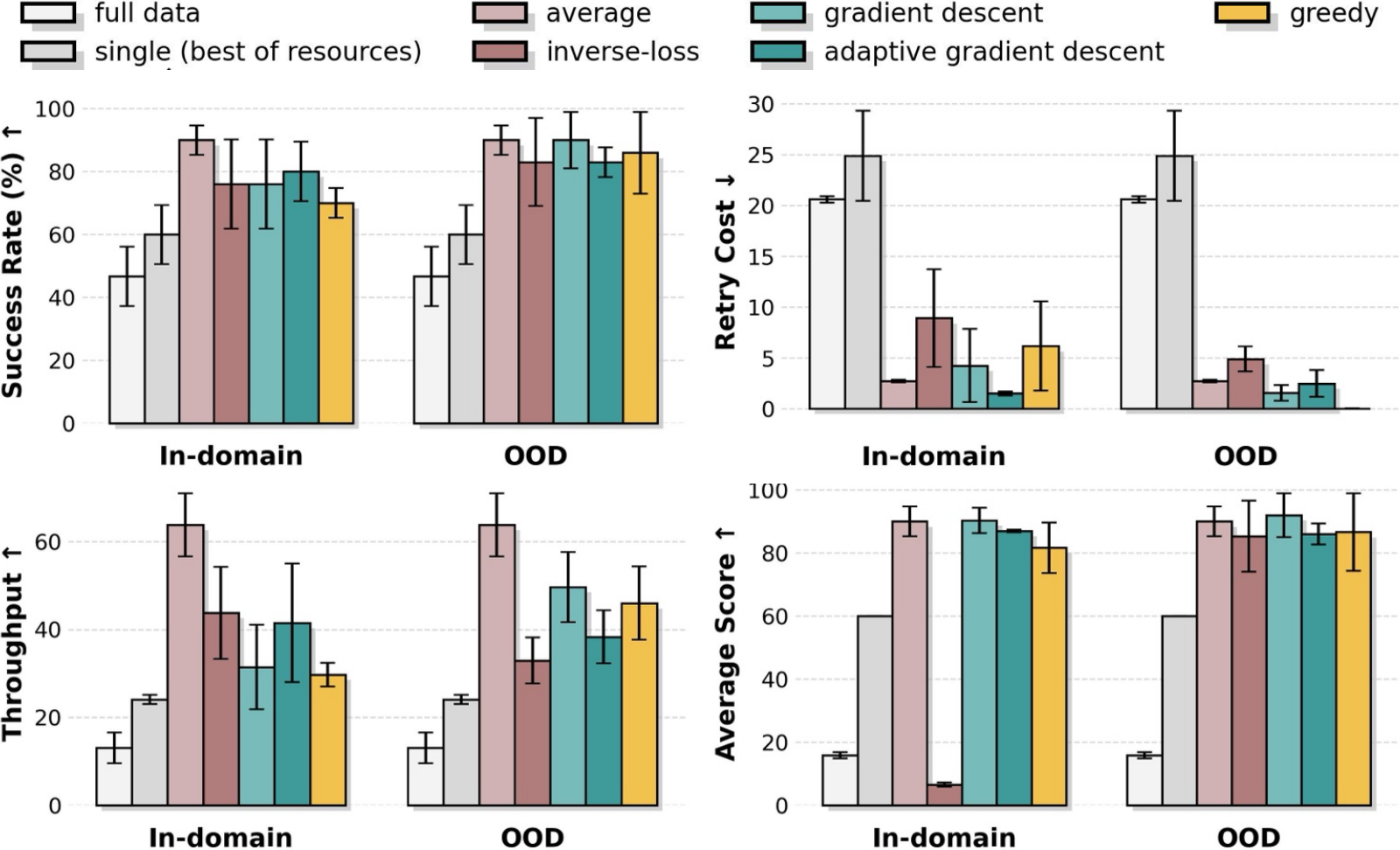}
  }
  \caption{\textbf{Ablations of MA on Tasks A and B.} Consistent with Task C, all MA variants outperform single-best and full-data baselines in success rate and throughput across both tasks. Notably in Task B, the retry cost for all MA variants decreases significantly. These performance gains and stability improvements are maintained in OOD validation settings.}
  \label{fig:supp_exp_soup_both}
  \vspace{-1em}
\end{figure}

\subsection{More Ablation}
\label{sec:supp_exp_tables}

\begin{figure}[t]
  \centering
  \includegraphics[width=0.6\columnwidth]{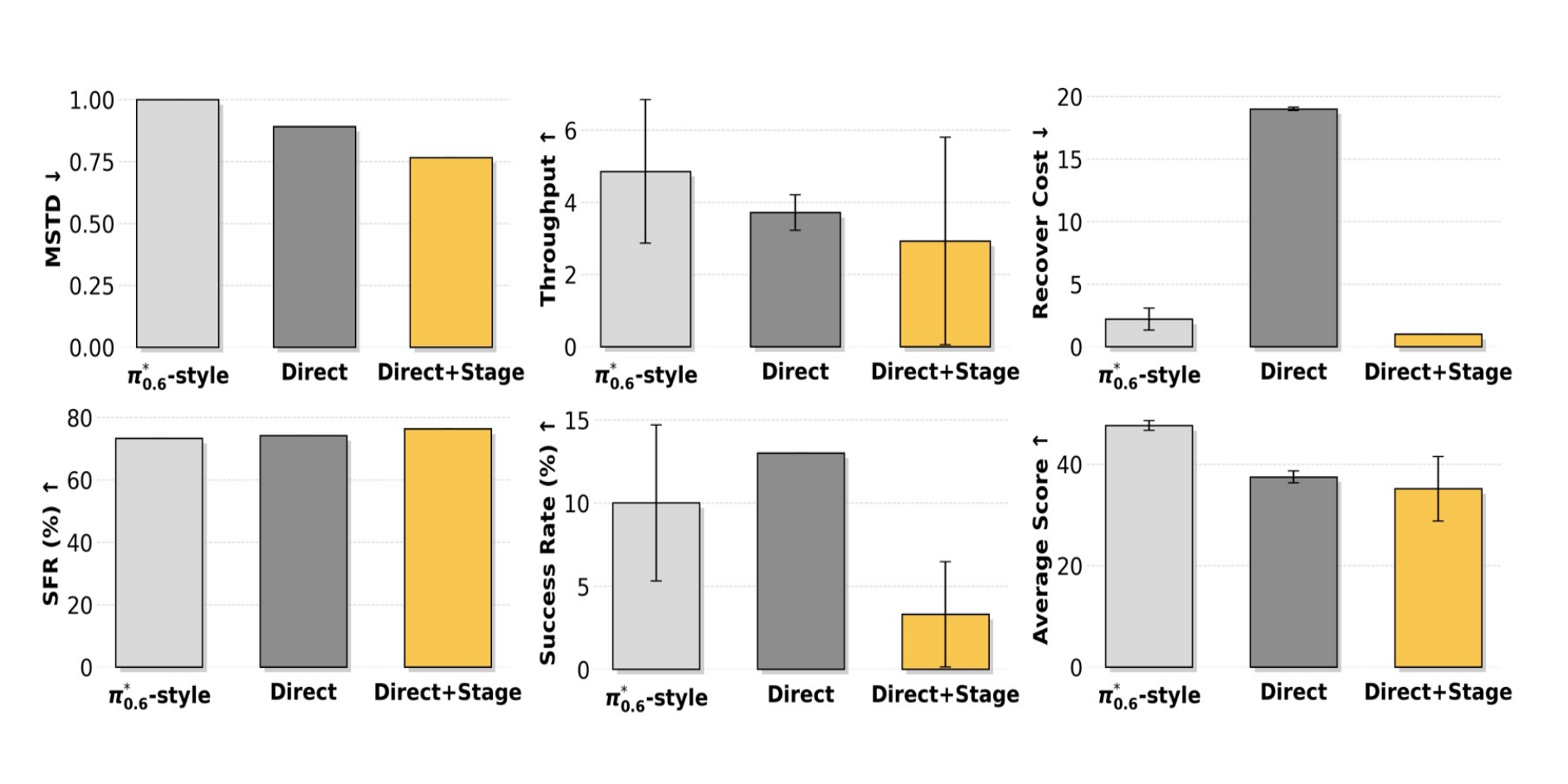}
    \vspace{-1em}
  \caption{\textbf{Ablations of SA on Task C.} SA demonstrates better numerical stability than $\pi^*_{0.6}$-style advantage baseline in terms of MSTD and SFR. Even though Direct + Stage variant sees a performance drop on this non-staged task, the Direct advantage method alone still consistently outperforms $\pi^*_{0.6}$-style.}
  \label{fig:supp_exp_advantage_hang}
\end{figure}

\begin{figure}[t]
  \centering
  \subfloat[\textbf{Absolute Joint}
    \label{fig:supp_exp_consistency_ff_abs}]{
    \includegraphics[width=0.48\columnwidth]{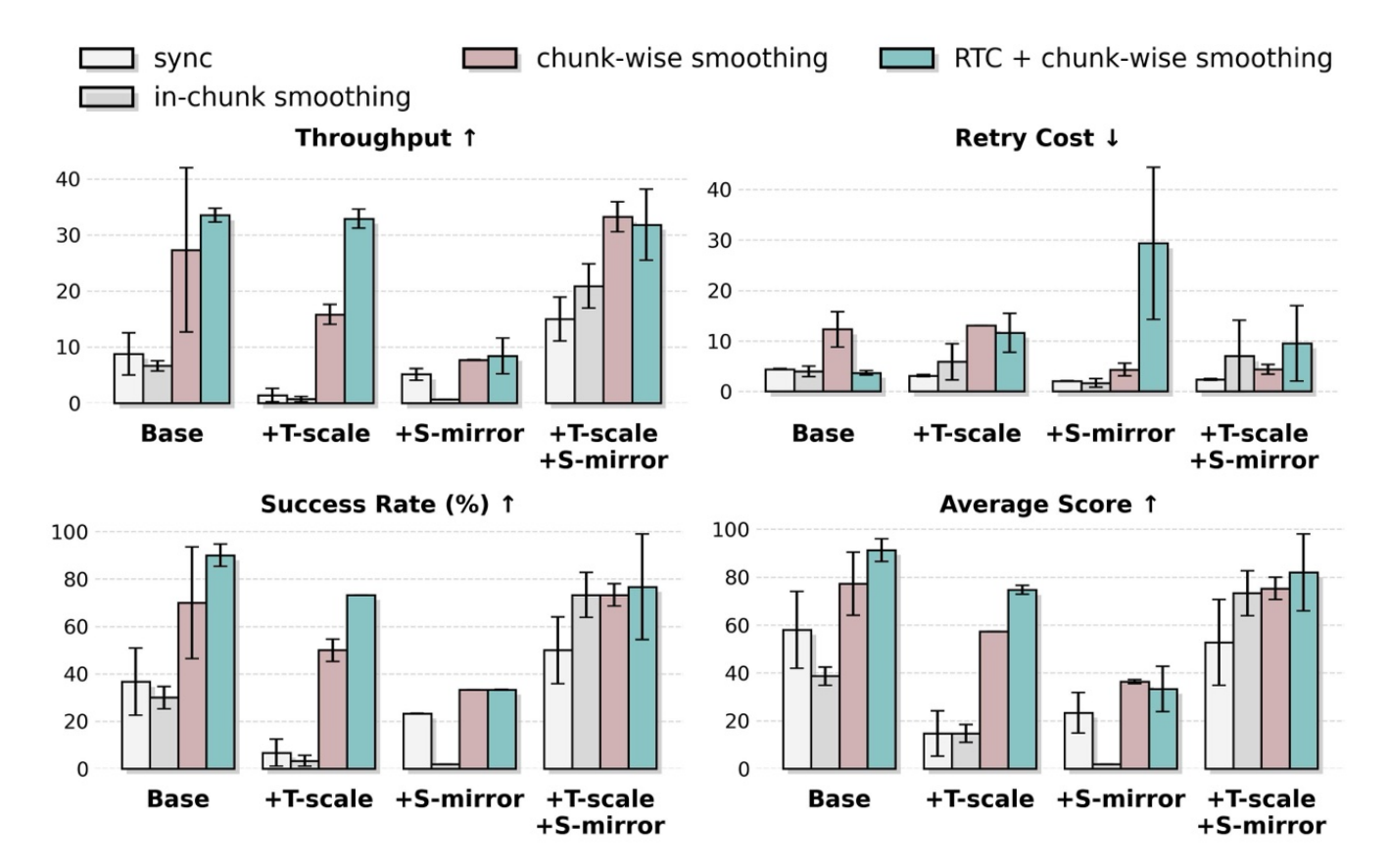}
  }
  \hfill
  \subfloat[\textbf{Delta Joint}
  \label{fig:supp_exp_consistency_ff_delta}]{
    \includegraphics[width=0.48\columnwidth]{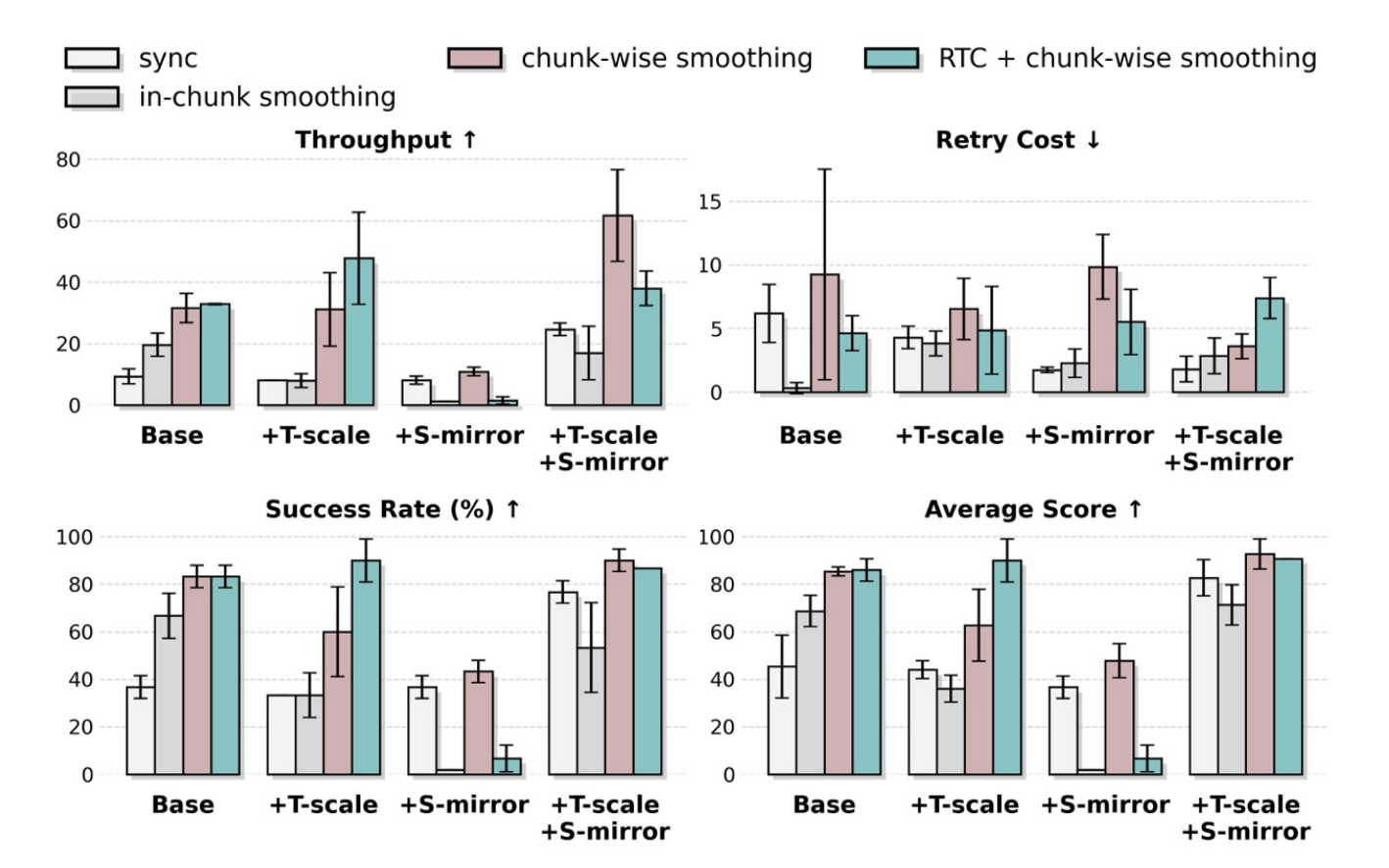}
  }
  \caption{\textbf{Ablations of control strategies and spatio-temporal augmentation on Task A with different action representations.} The figure compares performance using \textbf{Absolute Joint} (left) and \textbf{Delta Joint} (right) control. Across both action representations, we observe consistent trends: temporal chunk-wise smoothing significantly boosts inference performance, while combining it with RTC further improves task success rates across augmentation settings.}
  \label{fig:supp_exp_consistency_a}
  \vspace{-1em}
\end{figure}

\subsubsection{MA ablation on other tasks}

Figure~\ref{fig:supp_exp_soup_both} details the performance of Model Arithmetic (MA) on Tasks A and B. 
The results corroborate our primary finding: in most cases, MA consistently outperforms both the single-best candidate and the full-data baseline. 
However, the correlation regarding validation loss selection is less consistent. 
We attribute this discrepancy to the variable quality of the Out-Of-Distribution (OOD) data. 
We hypothesize that validation loss is a reliable proxy for candidate selection only when the OOD data distribution sufficiently approximates $P_\text{test}$.

\subsubsection{SA ablation on other tasks}

Figure~\ref{fig:supp_exp_advantage_hang} presents the ablation study of Stage Advantage (SA) on Task C. 
While numerical stability metrics mirror the improvements observed in Tasks A and B, this stability does not translate equivalently to final task performance. 
We hypothesize that this discrepancy stems from the $\pi_{0.5}$ pre-training distribution; the absence of hanging-specific priors likely limits the model's high-level planning capabilities, rendering it unable to fully exploit the improved numerical stability.

\subsubsection{TDA ablation on other tasks}

\begin{figure}[t]
  \centering
  \subfloat[\textbf{Absolute Joint}
    \label{fig:supp_exp_consistency_demoB_abs}]{
    \includegraphics[width=0.48\columnwidth]{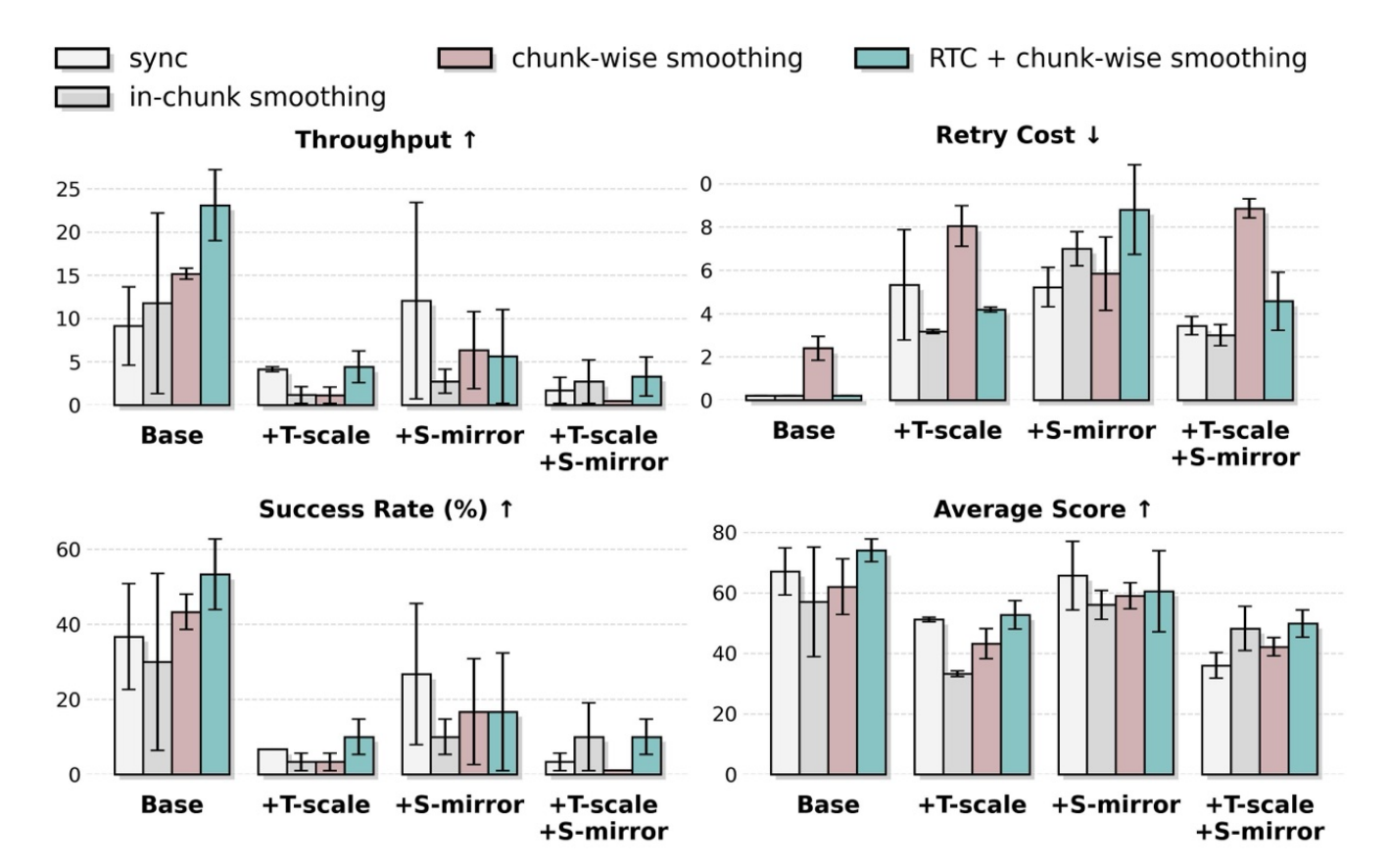}
  }
  \hfill
  \subfloat[\textbf{Delta Joint}
  \label{fig:supp_exp_consistency_demoB_delta}]{
    \includegraphics[width=0.48\columnwidth]{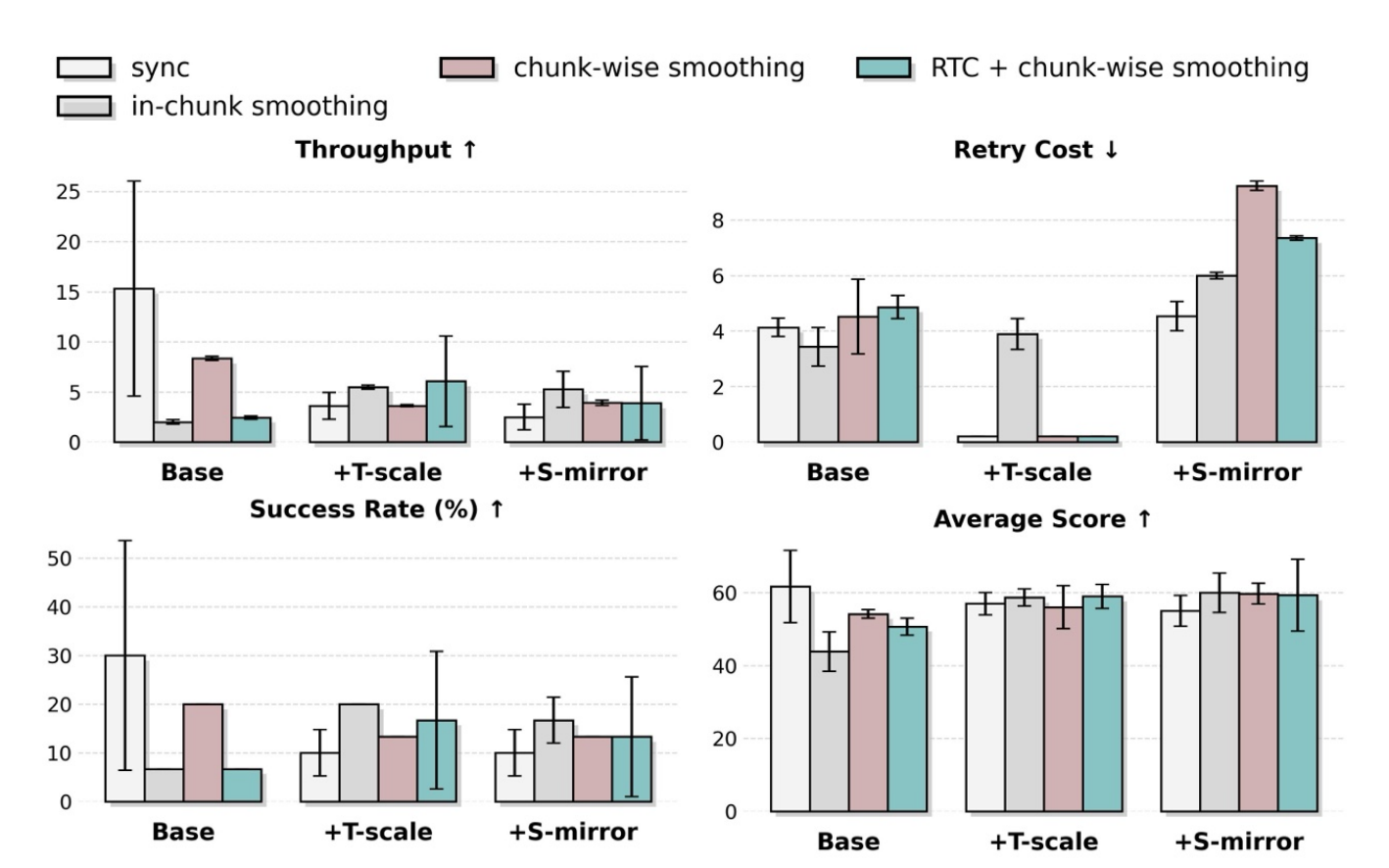}
  }
  \caption{\textbf{Ablations of control strategies and spatio-temporal augmentation on Task C with different action representations.} Since Task C is a fine-grained task, overall SR is relatively low for both absolute and delta joint action control. However, temporal chunk-wise smoothing consistently outperforms these non-smoothing baselines.}
  \label{fig:supp_exp_consistency_c}
  \vspace{-1em}
\end{figure}

\begin{figure}[ht]
  \centering
  \subfloat[\textbf{Absolute Joint}
    \label{fig:supp_exp_consistency_demoA_abs}]{
    \includegraphics[width=0.48\columnwidth]{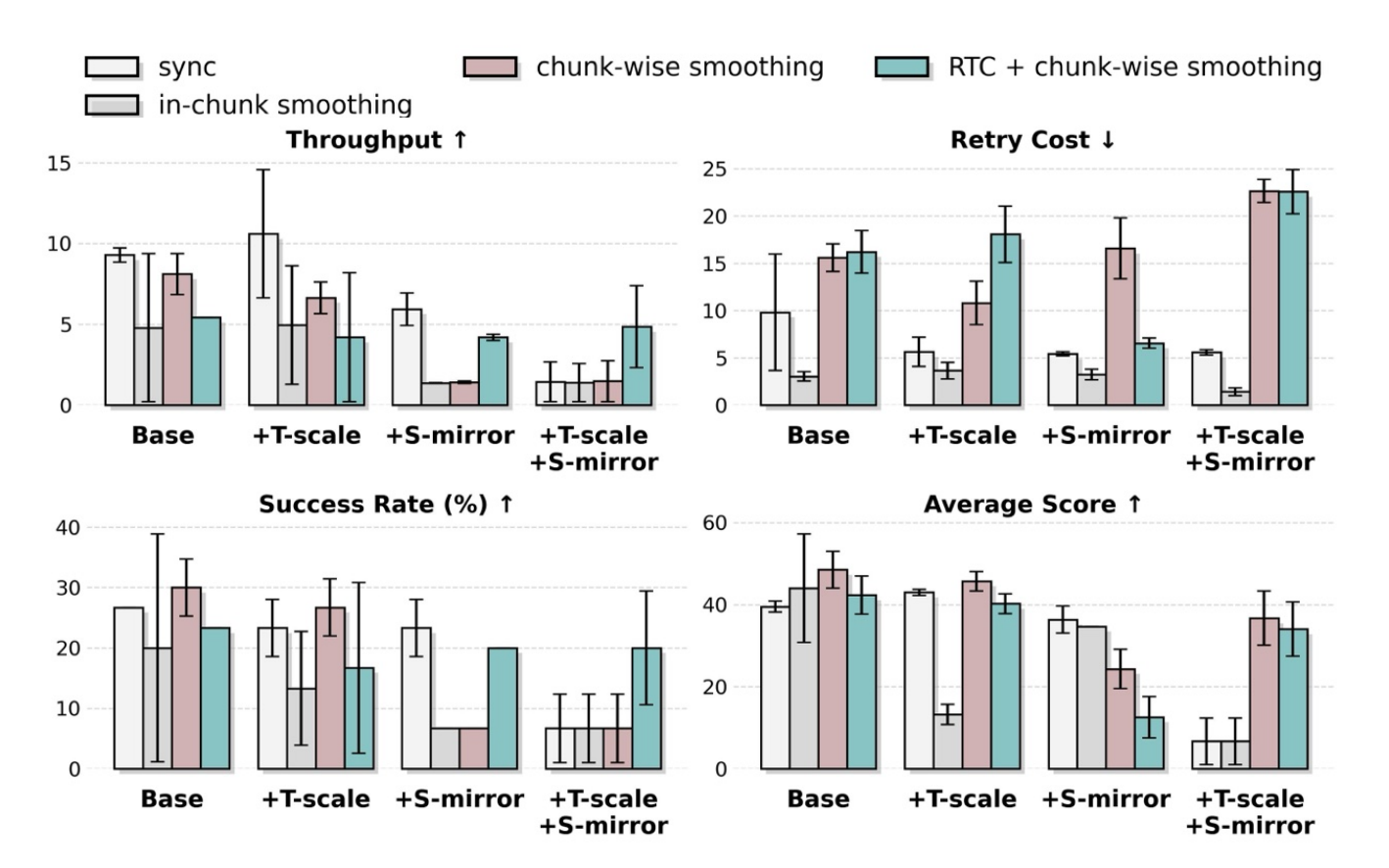}
  }
  \hfill
  \subfloat[\textbf{Delta Joint}
  \label{fig:supp_exp_consistency_demoA_delta}]{
    \includegraphics[width=0.48\columnwidth]{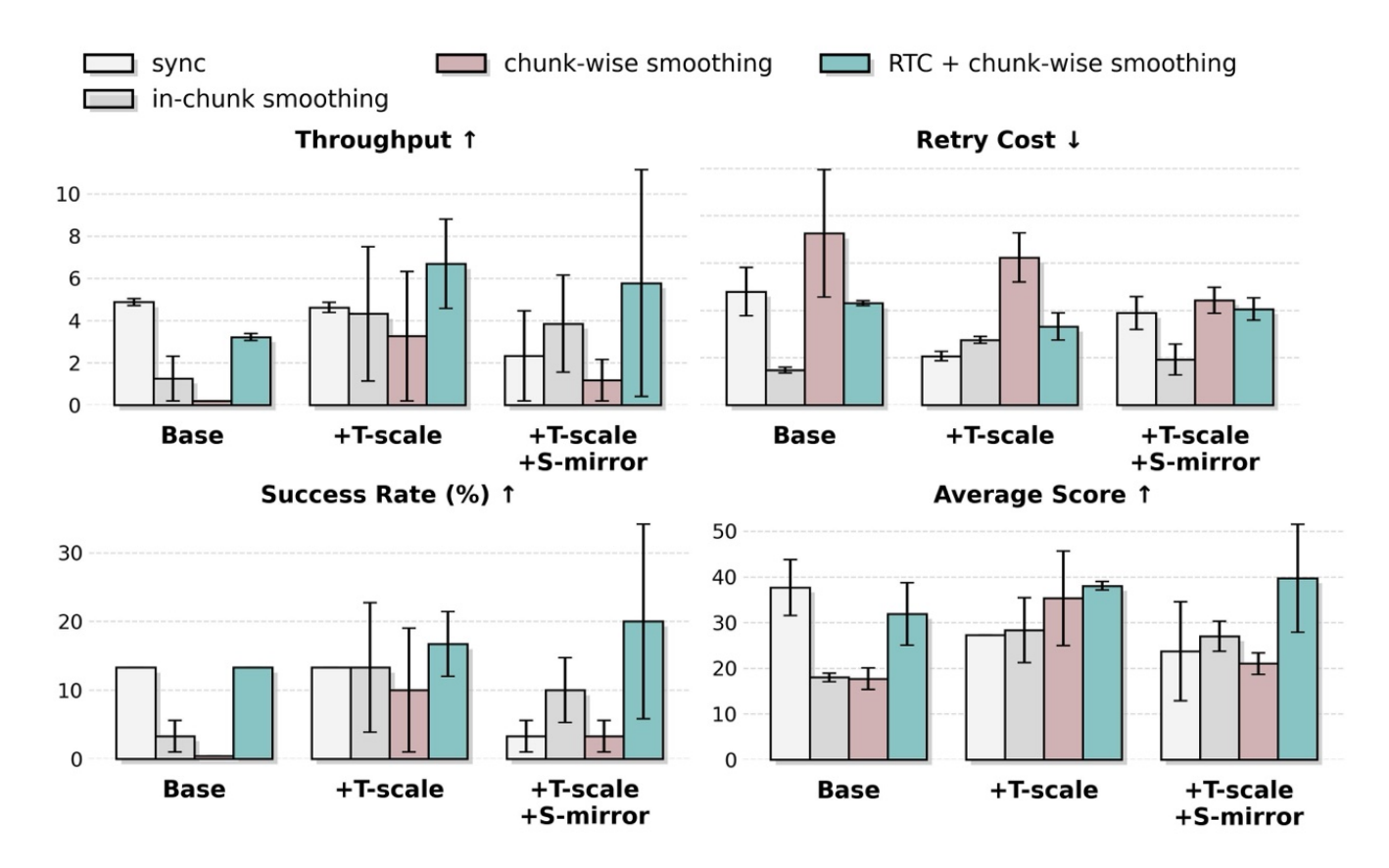}
  }
  \caption{\textbf{Ablations of control strategies and spatio-temporal augmentation on Task B with different action representations.}While delta joint control underperforms absolute joint control on this task, temporal chunk-wise smoothing consistently improves inference performance across both representations and augmentation settings.}
  \label{fig:supp_exp_consistency_b}
  \vspace{-1em}
\end{figure}

We evaluate the impact of absolute versus delta joint state prediction on performance, as detailed in Figure~\ref{fig:supp_exp_consistency_a} (Task A), Figure~\ref{fig:supp_exp_consistency_b} (Task B), and Figure~\ref{fig:supp_exp_consistency_c} (Task C). 
Extensive experiments validate the efficacy of our temporal chunk-wise smoothing across most of the settings.
Notably, Task C exhibits heightened sensitivity to action parameterization (absolute vs. delta) compared to Tasks A and B. 
We attribute this to the high-precision dexterity required for the hanger insertion phase.

\subsection{Failure case analysis and qualitative result}
\label{sec:supp_failure}

To answer \textbf{Q6}, we conducted a comprehensive qualitative analysis by visualizing and categorizing all failure instances recorded during Task A evaluation. The flattening stage proves to be the bottleneck of Task A, accounting for the majority of policy failures due to its high complexity. Consequently, DAgger and Heuristic DAgger are particularly effective here, as they naturally prioritize collecting recovery data in this critical phase where the robot is most prone to errors.

\subsection{Data ethics and license}
\label{sec:supp_data_ethics}
We will release the full code and data in the future.
Data is under CC BY-NC-SA 4.0 license and the code is under Apache-2.0 license.

\subsection{Contributions} 
\label{sec:supp_contributions}

All the names are ordered alphabetically by their first name in the first page. Below are the detailed taskforce synergy.

\smallskip

\noindent\textbf{Project Lead}. Chonghao Sima, Li Chen

\noindent\textbf{Core Members}. Checheng Yu, Chonghao Sima, Jin Chen, Kaiyang Wu, Lirui Zhao, Modi Shi, Yibo Yuan

\noindent\textbf{Model Arithmetic}. Chonghao Sima, Modi Shi

\noindent\textbf{Stage Advantage}. Lirui Zhao

\noindent\textbf{Train-Deploy-Alignment}. Checheng Yu, Chonghao Sima, Kaiyang Wu, Yibo Yuan

\noindent\textbf{Data Collection}. Checheng Yu, Chonghao Sima, Gangcheng Jiang, Kaiyang Wu, Lirui Zhao

\noindent\textbf{Deployment and Hardware Maintainance}. Checheng Yu, Chonghao Sima, Gangcheng Jiang, Jin Chen, Kaiyang Wu, Shijia Peng

\noindent\textbf{Writing and Illustration}. Checheng Yu, Chonghao Sima, Hongyang Li, Kaiyang Wu, Li Chen, Lirui Zhao, Modi Shi, Qingwen Bu, Tianyu Li, Yibo Yuan

\noindent\textbf{Discussion and Feedback}. Hai Zhang, Haoguang Mai, Huijie Wang, Ping Luo, Shijia Peng

\noindent\textbf{Resource and Academic Supervision}. Hongyang Li, Ping Luo

\end{document}